%% file: main.tex
\documentclass[10pt, conference, compsocconf]{IEEEtran}
\IEEEoverridecommandlockouts

\usepackage[dvipdfmx]{graphicx}
\usepackage{cite}
\usepackage{amsmath,amssymb,amsfonts}
\usepackage{algorithmic}
\usepackage{soul}
\usepackage{textcomp}
\usepackage{xcolor}
\usepackage{subfigure}
\usepackage{textcomp}
\usepackage{textgreek}
\usepackage{url}

\def\BibTeX{{\rm B\kern-.05em{\sc i\kern-.025em b}\kern-.08em
    T\kern-.1667em\lower.7ex\hbox{E}\kern-.125emX}}

\makeatletter
\newcommand{\linebreakand}{%
  \end{@IEEEauthorhalign}
  \hfill\mbox{}\par
  \begin{@IEEEauthorhalign}
}    

\begin{document}

\title{Privacy-preserving Pedestrian Tracking \\ using Distributed 3D LiDARs}

\author{\IEEEauthorblockN{Masakazu Ohno}
\IEEEauthorblockA{\textit{Osaka University} \\
Osaka, Japan \\
m-ohno@ist.osaka-u.ac.jp}
\and
\IEEEauthorblockN{Riki Ukyo}
\IEEEauthorblockA{\textit{Osaka University} \\
Osaka, Japan \\
r-ukyoh@ist.osaka-u.ac.jp}
\and
\IEEEauthorblockN{Tatsuya Amano}
\IEEEauthorblockA{\textit{Osaka University} \\
Osaka, Japan \\
t-amano@ist.osaka-u.ac.jp}

 \linebreakand
 \and
\IEEEauthorblockN{Hamada Rizk}
\IEEEauthorblockA{\textit{Osaka University,
Osaka, Japan} \\
\textit{Tanta University, Tanta, Egypt}\\
hamada\_rizk@f-eng.tanta.edu.eg}
\and
\IEEEauthorblockN{Hirozumi Yamaguchi}
\IEEEauthorblockA{\textit{Osaka University} \\
Osaka, Japan \\
h-yamagu@ist.osaka-u.ac.jp}
}

\maketitle

\input{definition}

\begin{abstract}
The growing demand for intelligent environments unleashes an extraordinary cycle of privacy-aware applications that makes individuals' life more comfortable and safe. Examples of these applications include pedestrian tracking systems in large areas. Although the ubiquity of camera-based systems, they are not a preferable solution due to the vulnerability of leaking the privacy of pedestrians.
In this paper, we introduce a novel privacy-preserving system for pedestrian tracking in smart environments using multiple distributed LiDARs of non-overlapping views. The system is designed to leverage LiDAR devices to track pedestrians in partially covered areas due to practical constraints, e.g., occlusion or cost. Therefore, the system uses the point cloud captured by different LiDARs to extract discriminative features that are used to train a metric learning model for pedestrian matching purposes. To boost the system's robustness, we leverage a probabilistic approach to model and adapt the dynamic mobility patterns of individuals and thus connect their sub-trajectories.
We deployed the system in a large-scale testbed with 70 colorless LiDARs and conducted three different experiments. The evaluation result at the entrance hall confirms the system's ability to accurately track the pedestrians with a 0.98 F-measure even with zero-covered areas. This result highlights the promise of the proposed system as the next generation of privacy-preserving tracking means in smart environments.
\end{abstract}


% ===
% Introduction
% ===
\section{Introduction}

Many systems have been proposed to detect and track people in large areas, especially using RGB cameras (the nowadays commodity devices for such a purpose). 
Those camera-based tracking systems usually use multiple RGB cameras to track people (referred to as Multi-Camera Tracking (MCT)) %\cite{dendorfer2021motchallenge, tang2019moana}.
\cite{dendorfer2021motchallenge}.
Since person Re-ID (re-identification) among multiple cameras is the primary issue in MCT, 
recent MCT approaches rely on the power of deep learning techniques to get features from RGB cameras.
These features are used to identify each person who appears in different camera scenes to enable tracking of the target person. 
However, these features are often bio-metric features of the bystander, e.g., face, skin color, gender, age, and body shape. 
Deep neural networks can encode the image features into latent space, which is safer (\textit{i.e.} unable to recover the original data) as the structure of the networks becomes complicated. 
Nevertheless,  there have been many attacks that attempt to leak the privacy of users (along with their images) involved in such systems  methods like membership inference attacks \cite{shokri2017membership}
and model inversion attacks \cite{fredrikson2015model}.

Recently,  3D Light Detection and Ranging sensors (3D LiDARs, or simply LiDARs) have attracted more attention in terms of the 
balance between the privacy-preserving features and the capability of spatial sensing.  
LiDARs only acquire distances to the nearest objects in each 3D direction in FoV (Field of View),  and the distance error is usually in the range of $10^{-2}\sim 10^{-1}$ meters with a coverage range of up to $100$ meters. LiDARs generate colorless 3D point clouds,
which is more privacy-reserving compared to camera-based systems.
Moreover, 3D LiDARs enjoy accurate ranging characteristics making their pedestrian tracking accuracy generally higher than the camera-based counterparts \cite{shackleton2010tracking,alvarez2019people}.

This gives rise to the development of large-scale pedestrian tracking systems with Multi-LiDARs (MLT) distributed over the target area. However, full coverage of large areas with LiDARs (or even cameras) may not always be possible due to the associated cost, the lack of power supplies, and/or occlusion (LoS constraints)\footnote{one common issue in indoor tracking using cameras/LiDARs is how to obtain clear views to track pedestrians with a lot of obstacles like ad signs, plants, etc.}. Therefore, pedestrian tracking with multi-LiDARs in partially covered areas is a more challenging problem. In other words, the problem is to track pedestrian(s) captured with one LiDAR and re-identify them with a colorless point cloud of other LiDARs given a non-overlapping field of views (FoVs) and a non-covered area in-between. 

In this paper, we propose a privacy-preserving pedestrian tracking system using distributed 3D LiDARs of non-overlapping views. 
Specifically, the proposed system attempts to find the correspondence between multiple pedestrians' sub-trajectories obtained by each LiDAR to estimate the whole trajectory of each person in the target area. 
Towards this end, the proposed system identifies pedestrians and recognizes their sub-trajectories based on two criteria: the similarity of the point cloud signature of each pedestrian and the Spatio-temporal characteristics of the pedestrian sub-trajectories.
The former is achieved by employing the Fisher Vector approach  \cite{ben20173d} to 
extract discriminative fixed-size features representing the shape and behavior of each person given her point cloud. 
The system then trains a deep-metric learning model to learn the dissimilarity between the features of different persons.
The second criterion leverages sub-trajectory start/end points to learn possible point transitions of pedestrians. This is done by defining the probability distribution of traveling time and mobility patterns and updating these distributions using the Bayesian approach.  

To demonstrate the usefulness of the proposed approach, we deployed the system in a large testbed of six floors building equipped with 70 LiDARs and conducted three different experiments. 
The evaluation results of the system on 32 pedestrians confirm its efficacy in achieving a consistently high matching accuracy of the pedestrian trajectories with 0.98 F-measure. This result is achieved with only colorless sparse 3D LiDARs that ensure the privacy of pedestrians.

To summarize, our contributions are three-fold.
(i) \textit{Uniqueness of the problem.} We tackle a new problem to obtain complete pedestrian trajectories from sub-trajectories which are captured by distributed 3D LiDARs. As far as we investigate, no other research has been done for this problem.
(ii) \textit{Novelty of the approach.} 
    Unlike multi-camera multi-object tracking, the LiDAR-based sub-trajectories are more accurate, but we have less clue to connect those segmented ones in terms of person re-identification. 
    To address the issue, 
    we design a unique algorithm to find the most likely matching among them 
    taking point cloud-specific features as input. The probability distribution functions are updated based on the Bayesian updating system.
(iii) \textit{Evaluation using the real data.}  We conducted several experiments using the real LiDAR data obtained in our large-scale testbed. The testbed consists of 70 LiDARs installed over 6 floors of the 7-story building in our university campus. The performance of the method has been validated through the dataset.

% ===
% Related Work
% ===
\section{Related Work}
Multi-Object Tracking (MOT) has extensively studied  to track persons or vehicles \cite{dendorfer2021motchallenge}. Solution for this problem can be, in general, categorized into \textit{single camera approaches} and \textit{multiple camera approaches}.

\subsection{Multi-Object Tracking (MOT) with Single Camera / LiDAR}

Tracking a person or vehicle using a single camera has been well investigated due to the availability of  several public datasets, e.g., MOT Challenge\cite{MOTChallenge20} and KITTI dataset\cite{geiger2012we}. 
The MOT Challenge dataset consists of videos captured by surveillance cameras enabling tracking methods in various scenarios, such as tracking in congested areas.
The KITTI dataset includes 3D point cloud data of pedestrians acquired using LiDARs on top of passing vehicles.
DeepSORT\cite{wojke2017simple} is a typical MOT system that uses the Yolo  object detection method to detect target objects in a given frame. Then, it leverages Kalman filter to track the moving objects in consecutive frames.
3D vehicle detection and tracking from monocular videos has been proposed in \cite{Hu_2019_ICCV}. It can estimate  3D bounding boxes surrounding each object in a sequence of 2D images, using a Deep Neural Network (DNN).

On the other hand, a lot of efforts have been dedicated to 3D tracking using 3D cameras or LiDARs. RGB images and 3D point clouds have been leveraged together for tracking pedestrians in  \cite{8462884,9341635,8818343},
while 3D point clouds have been adopted alone in \cite{9352500,Simon_2019_CVPR_Workshops,9341164,Spinello_Arras_Triebel_Siegwart_2010,5649769}.
An advantage of using 3D point clouds alone is the resilience to varying colors and brightness, which usually affect cameras. Additionally, the privacy concerns associated with surveillance cameras do not apply to LiDARs. However, leveraging LiDARs makes MOT challenging as colors are the most critical signature to detect/identify persons. 
To cope with the issue, self-designed features are often incorporated to perform 3D MOT using 3D point clouds \cite{6728343,7237897}.
We have also used our own tracking system using a single 3D LiDAR. 

\textit{Different from these apporaches, the proposed system is designed to achieve accurate pedestrian tracking  using multiple, distributed 3D LiDARs. }

\subsection{Multi-Camera Multi-Object Tracking with Person Re-Identification}
Multi-Camera Multi-Object Tracking (Multi-Camera MOT)
refers multiple object tacking with multiple distributed cameras which has been investigated for specific cases in \cite{ristani2018features}, \cite{tang2019cityflow}, \cite{hsu2019multi}.
While identifying a person detected by one camera and using a different camera at a different location and timing (called \textit{person re-identification} in multi-camera MOT) has been proposed  \cite{ge2020self}.
In \cite{Ikegame2005}, the similarity between pedestrians is calculated using the clothes' colors  in RGB images and the traveling time and distance between the different cameras. Then, the matching is carried out using the Hungarian method\cite{munkres1957algorithms,bourgeois1971extension}. Additionally, different types of camera are fused  for  person re-identification \cite{mogelmose2013tri,wang2019rgb}. The system in \cite{mogelmose2013tri} combines a RGB-D camera with a temperature sensor, while RGB and infrared cameras are used in \cite{wang2019rgb}.
In \cite{mogelmose2013tri}, RGB-D camera with  depth data have been utilized to extract the  pedestrians' skeletal information. Also, recent studies, e.g., \cite{7864367,10.1145/3506708}, tends to enhance the capability of RGB-D based recognition and person re-identification. 
On the other hand, the system in \cite{xu2021review} estimates joints of human bodies from a 3D point cloud. However, this work cannot be directly adopted for person re-identification as the human pose does not contain enough information to distinguish persons.
Although the presence of datasets for camera-based person re-identification, e.g.,
Market-1501\cite{zheng2015scalable} and Motion Analysis and Re-identification Set (MARS)\cite{zheng2016mars}, no similar datasets are available using multiple distributed  3D LiDARs.

\textit{To the best of our knowledge, this is the first work that leverages the privacy-preserving 3D point clouds captured by multiple distributed LiDARs for person re-identification. Additionally, the proposed system handle the challenges associated with processing 3D point cloud, such as unordered, unstructured, and varying size point clouds.}

\subsection{Trajectory Prediction}

Trajectory prediction\cite{TraPlan,TrajectoryPrediction} has been studied to extract patterns from moving trajectories.
The method in \cite{TraPlan} proposes a trajectory prediction model for traffic networks based on past movement patterns.
By dividing trajectories into clusters and discovering frequently occurring trajectory patterns, future movement trajectories can be predicted with high accuracy.
The authors of \cite{TrajectoryPrediction} also investigate trajectory prediction, but they use sparse coding\cite{Chen2016} to represent each trajectory as a combination of predefined trajectory patterns, which enables the prediction of subsequent trajectories for each new trajectory.
They also propose a similarity-based model fusion algorithm that allows agents to update their knowledge by communicating the data they have learned with each other.

\textit{On the contrary, this paper focuses on the problem of finding pedestrians trajectories from a given set of their sub-trajectories, 
the context of our work is entirely different.}

% ===
% Motivation for using LiDARs
% ===
\section{Motivation for Using LiDARs}
In this section, we motivate the adoption of LiDARs as the core technology for the proposed system.
LiDAR is emerging as a powerful enabler of the next generation of smart and safe environments~\cite{gain2020,10.1145/3539659}.
LiDARs can provide long-range, real-time, centimeter-level distance measurements of surrounding objects in all lighting conditions.  

\textbf{Privacy:} LiDAR provides a key advantage over camera-based systems – privacy protection. With increased concerns that facial recognition technology can be used for general surveillance, the U.S. Congress discusses legislation that seeks to ban the use of camera-based human identification and other biometric surveillance technology by federal law enforcement agencies.
 Thus, nowadays, industry has seen leading tech vendors stepping away from their own camera-based facial recognition technologies, as reported in Forbes~\cite{forbes1}\footnote{IBM plans to leave the facial recognition business, Amazon is placing a one-year hold on police departments using its facial recognition technology, and Microsoft is waiting on federal legislation before the company starts selling its comparable technology to law enforcement.}.  In the U.S. Congress, there is legislation that seeks to ban the use of camera-based facial recognition and other biometric surveillance technology by federal law enforcement agencies. 
 On the other hand, a lot of effort has been devoted to  preventing/reducing camera's capabilities from obtaining detailed visual data (private information) by equipping cameras with additional hardware/software\cite{Hinojosa_2021_ICCV,7778202}. 
In contrast, LiDAR, by definition,  captures only point cloud representation of the scene, from which humans' biometric features, such as facial characteristics, hair and skin color, or even clothes, cannot be identified.

\textbf{Cost: }One strength commonly associated with cameras when compared to LiDARs is cost. However, when the system design necessitates optimal levels of privacy, coverage, and varying lighting conditions, the assumed advantage of camera-centric approaches diminishes greatly. A single LiDAR sensor can typically cover roughly four times the area of one camera, significantly decreasing the costs and logistics of installation. Moreover, nowadays, LiDARs are becoming as cheap as only 80\$~\cite{magikeye} and reliable in different applications\cite{9767322,9826758,katayama,9941552,10.1145/3495243.3558744}.

\textbf{Setup: }A LiDAR-based solution has setup efficiency and simplicity benefits over camera-based approaches. Using high-quality LiDAR, which generates dense point clouds at longer ranges, enables reliable tracking at scale with fewer devices. Additionally, LiDAR data is much faster and simpler to process, requiring less computing power within a system compared to cameras.

% ===
% System Architecture and Problem Definition
% ===
\section{System Architecture and Problem Definition}
\label{sec:tracking}

\subsection{Obtaining Sub-Trajectories by Each LiDAR} 

Each LiDAR can capture a part of the target 3D space, and 
in each frame (\textit{i.e.} one scan of the space) from the data stream from the LiDAR, 
a 3D point cloud (or simply point cloud) is obtained.
The 3D space where the point clouds are generated by LiDAR $i$ is called \textit{scan space} of LiDAR $i$ and denoted as $\sthree{i}$. 
Then a background subtraction method is applied to that point cloud to extract moving objects. 
The extracted point cloud is called the foreground point cloud, and segmentation is applied to find each person in the foreground point cloud.
We use the Voxel Grid Filter\cite {5980567} and apply downsampling to the foreground point cloud to convert each voxel grid cell into a single virtual point.
Then, a clustering algorithm is applied to the foreground point cloud to segment it into \textit{human segments}.
We remove the Z-axis when the clustering is applied to reduce processing overhead and employ a DBSCAN-based clustering to obtain human segments. 

\begin{figure}[t]
 \vspace{-0.5cm}
    \centering
    \includegraphics[width=\linewidth]{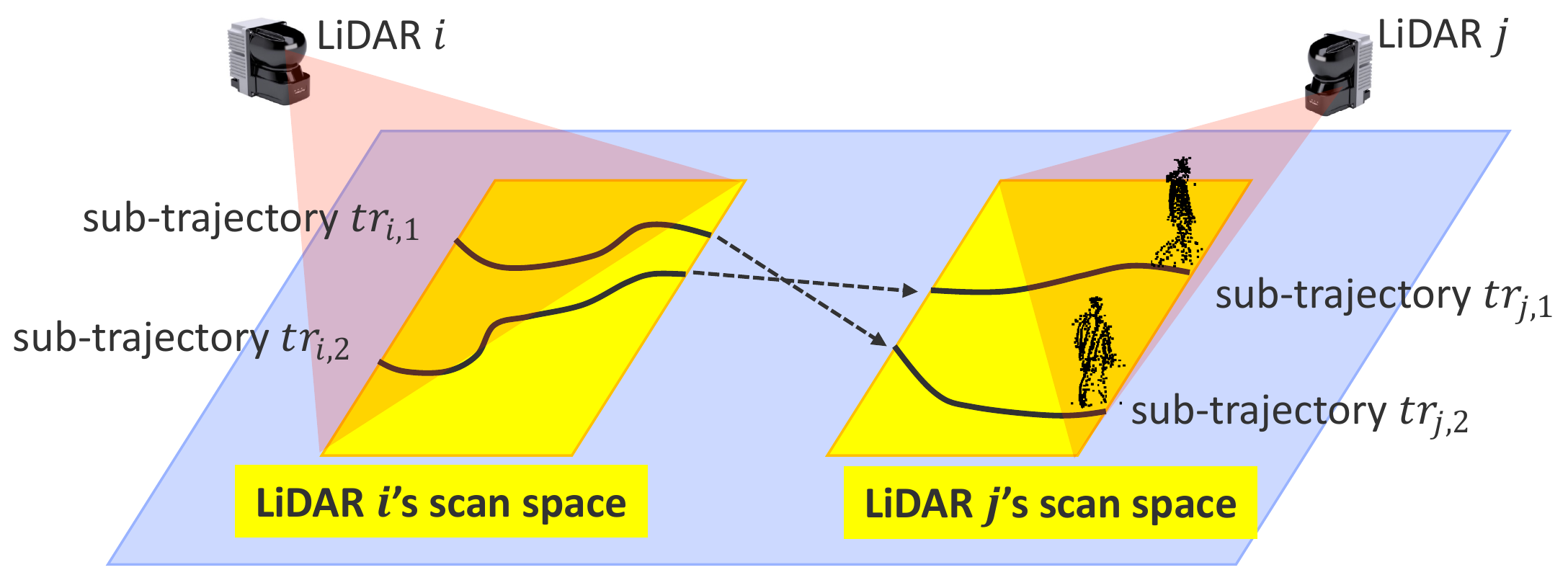}
    \vspace{-0.6cm}
    \caption{Sub-trajectories by Distributed LiDARs}
    \label{fig:problem_image}
     % \vspace{-0.5cm}
\end{figure}

A \textit{sub-trajectory} refers to a two-dimensional trajectory obtained as a temporal sequence of the $(x,y)$-coordinates of human segments, which corresponds to one person's walking trajectory in a LiDAR's scan space $\sthree{i}$ (Fig. \ref{fig:problem_image}).
We note that the prefix ``sub-'' indicates that the sub-trajectory represents only a part of the whole trajectory of one person.
For simplicity, the 2D area where sub-trajectories can be obtained by LiDAR $i$ is called \textit{trajectory area} of LiDAR $i$ and denoted as $\stwo{i}$. Generally, $\stwo{i}$ is the projection of $\sthree{i}$ onto the XY-plane.

We let $\trajset{[t, t']}{i}$ and $\segset{t}{i}$ denote the set of sub-trajectories obtained in a time window $[t, t']$ and the set of human segments at time $t$ by LiDAR $i$, respectively.
We also let $\traj{[t_a, t_b]}{i}{j}$ denote a sub-trajectory $j$ obtained by LiDAR $i$, which starts at time $t_a$ and ends at $t_b$.
We note that for any sub-trajectory 
$\traj{[t_a,t_b]}{i}{j} \in \trajset{[t, t']}{i}$, 
$t \leq t_a \leq t_b \leq t'$ holds.
$\traj{[t_a, t_b]}{i}{j}$ is contained in $\stwo{i}$ and their start and end points are on the boundary of $\stwo{i}$.

At time $t$, we obtain from LiDAR $i$ the set $\segset{t}{i}$ of human segments. 
We assume that the set $\trajset{[t-k, t-1]}{i}$ of sub-trajectories for $k-1$ time window ($k>1$) has been obtained.
Then we find the correspondence between a sub-trajectory $tr \in \trajset{[t-k, t-1]}{i}$, and a human segment $h \in \segset{t-1}{i}$. 
We can easily find this correspondence by applying our Kalman-filter-based tracking method \cite{9767224}, and we obtain an updated $\trajset{[t-k, t]}{i}$.
Finally, we define a temporal relation of two sub-trajectories, $\traj{[t_a, t_b]}{i}{u}$ and $\traj{[t_c, t_d]}{j}{v}$. 
We denote $\traj{[t_a, t_b]}{i}{u} < \traj{[t_c, t_d]}{j}{v}$ if and only if $t_b < t_c$ holds.  

% \section{Problem Definition and System Overview}

\subsection{Problem Definition}

We consider a time window, say $[\t{s}, \t{e}]$.
We assume that a set $\lidarset$ of LiDARs, their scan spaces and trajectory areas, and the map of the target area where those LiDARs are installed are given.
We also assume that a sub-trajectory set $TR = \bigcup_{i \in \lidarset} \trajset{[\t{s}, \t{e}]}{i}$ is obtained.

The \textit{trajectory estimation problem} in this paper is formulated as a problem to find a partition of $TR$, where each partition can form a \textit{sub-trajectory sequence}.
A sub-trajectory sequence is a temporal sequence of sub-trajectories satisfying the total order relation based on the temporal relation $<$. 
For example, a partition that contains sub-trajectories 
$tr_1 = \traj{[1, 3]}{i}{u}$, $tr_2 = \traj{[4, 6]}{j}{v}$ and $tr_3 = \traj{[10, 12]}{j}{r}$, satisfies $tr_1 < tr_2 < tr_3$, and can form the sub-trajectory sequence $tr_1;tr_2;tr_3$.

Once a set of partitions is found, for each partition and the map, we may estimate the path between the subsequent sub-trajectories using the corridors or pathways information contained in the map. 
However, due to space limitations, this is not in the scope of this paper. 

% 入力として以下が与えられる。

% \begin{itemize}
%     \item ある歩行者$i$のmフレーム分の点群の集合$P_i$の出現した歩行者n人分の集合$Q$
%     \item ある歩行者$i$のmフレーム分の軌跡の座標の集合$S_i$の出現した歩行者n人分の集合$T$
% \end{itemize}

% 1人1フレーム分の点群を$p\in P \subset Q$、座標を$s\in S \subset T$とする。
% 点群同士の類似度を求めるために距離関数distを作成する。2つの点群の集合$P_1,P_2$を入力として、2つの点群集合のコサイン類似度を返す。
% 観測するマップの情報はあらかじめ与えられており、観測範囲の境界には出入りがよくある区分に境界番号$b$が振られている。
% 歩行者$i$と歩行者$j$が同一人物と判定した場合、ペア$(i,j)$を集合$A$に追加する。
% 出力として以下を求める。
% \begin{itemize}
%     \item 同一人物と判定されたペアの集合$A$
% \end{itemize}

\if0
The inputs of our problem are the followings:
\begin{itemize}
    \item set $Q$ which contains point clouds' set $P_i$ obtained from a pedestrian $i$ for $m$ frames.
    \item set $T$ which contains coordinates of trajectories' set $S_i$ obtained from a pedestrian $i$ for $m$ frames.
\end{itemize}
$p\in P \subset Q$ is a point cloud and $s\in S \subset T$ is the coordinate for one frame of one pedestrian.
To find the affinity between the point groups, we create a metric function $dist$, which takes two sets of point groups $P_1,P_2$ as input and returns the cosine similarity between the two sets of point groups.
The information on the map to be observed is given in advance, and the boundaries of the observation area are assigned a boundary number $b$ to the segments that are frequently entered and exited.
If pedestrian $i$ and pedestrian $j$ are determined to be same person, pair $(i,j)$ is added to set $A$.
The output of the problem is the following value:
\begin{itemize}
    \item set$A$ of pairs determined to be the same person.
\end{itemize}
\fi

\subsection{Algorithm for Trajectory Estimation}

For a given $TR = \bigcup_{i \in \lidarset} \trajset{[\t{s}, \t{e}]}{i}$, our algorithm works as follows. 

We prepare two sets $V_1$ and $V_2$ of sub-trajectories and a set $E$ of sub-trajectory pairs, and all are initially empty. 
$V_1$ and $V_2$ correspond to the sets of those sub-trajectories whose end and start points are the connecting points, respectively.  
Then for every pair of sub-trajectories $tr_u, tr_v \in TR$, if $tr_u < tr_v$, we add $tr_u$, $tr_v$ and $(tr_u, tr_v)$ to $V_1$, $V_2$ and $E$, respectively. 
We also calculate the \textit{affinity value} ($\in [0,1]$) of the pair, which is defined and explained in Section \ref{sec:method}, and make an weight function $W: E \rightarrow [0,1]$. 

Finally, we obtain a weighted bipartite graph $G = (V_1 \cup V_2, E, W)$.
since $|V_1|=|V_2|$ holds according to the way to build $V_1$ and $V_2$, 
the problem is induced to find the optimal one-to-one matching of $V_1$ and $V_2$ over $E$.
This is equivalent to finding the subset $E'$ of $E$, which maximizes the total sum of the affinity values, as indicated in Eq. (\ref{eq:matching}).
\begin{equation}
E'= \arg \max_{E' \subseteq E} \;\; \Sigma_{e \in E'} W(e)
\label{eq:matching}
\end{equation}
For this problem, we can employ the Hungarian algorithm with $O(N^3)$ 
to find the optimal matching \cite{munkres1957algorithms}. 

It should be noted that the number of matching candidates, $V_1$ and $V_2$, can naturally be smaller if $\t{s}-\t{e}$ of $TR$ is smaller. 
This leads to the design of an online version of the matching algorithm. 
Specifically, we keep monitoring the sub-trajectories and updating $V_1$ and $V_2$, and once $|V_1|$ ($=V_2$) reaches a sufficient number, we can calculate the optimal matching and continue the procedure. 
The choice of time window size depends on the target applications and services.

\section{Sub-trajectory Affinity Calculation}
\label{sec:method}

In this section, we define an affinity value for each pair $(tr_u, tr_v)$ of sub-trajectories. To do so, we exploit the following three features, 
(i) similarity of two human segments (point clouds) from $tr_u$ and $tr_v$, respectively, 
(ii) statistical spatial feature (frequency of transitions) from the end point of $tr_u$ to the start point of  $tr_v$, 
and  
(iii) statistical temporal feature (traveling time) from the end point of $tr_u$ to the start point of  $tr_v$.  
The corresponding probabilities (likelihoods) are represented as $P_1$, $P_2$, and $P_3$, respectively, all of which range within $[0,1]$. 
The affinity value, denoted as $\aff{tr_1}{tr_2}$, is a multiplication of the above probabilities.
\begin{equation}
    \aff{tr_u}{tr_v} = P_1 \cdot P_2 \cdot P_3 
\end{equation}
In Sections \ref{subsec:human}, \ref{subsec:spatial} and \ref{subsec:temporal}, we explain how $P_1$, $P_2$ and $P_3$ are calculated, respectively. 
Besides, we will incorporate the Bayesian system to update the likelihood distributions of $P_2$ and $P_3$ as they are based on statistics, \textit{i.e.}, the prior distributions. We explain the update in Section 
\ref{subsec:beyesian}.

\subsection{Similarity of Human Segments}
\label{subsec:human}

We define the similarity of the two segments and calculate it to judge whether a pair of human segments (human point cloud segments) is from the same person or not.
Straightforward adoption of any learning-based similarity scheme is generally inadequate since the point cloud data is usually unordered and unstructured, and the number of points in a segment differs. 
Accordingly, we design Fisher Vector-based feature extraction and deep metric learning-based similarity calculation to tackle the problem.

\subsubsection{Feature Extraction}

We employ the Fisher Vector (FV) method to extract fixed-size representations of the input human segments. 
Specifically, FV computes the deviation of a 3D point cloud from the Gaussian Mixture Model (GMM). 
The intuition behind using FV for feature extraction is its ability to capture the spatial formation of 3D points in space, yielding discriminative signatures of human segments. 
This can be done by calculating the gradients of the sample’s log-likelihood with respect to the GMM model parameters (\textit{i.e.}, Gaussian weight, mean, and covariance).
The extracted feature representation of FV has a fixed-size independent of the number of points in a human segment. 
This advantage makes it easier to process variable-size human segments using a learning-based similarity technique.

Formally speaking, let $X_i = \{ \boldsymbol{p}_t \in \mathbb{R}^3, \; t=1,...,T\}$ be the set of 3D points of a human segment $i$, where $T$ denotes the number of points in a segment that dramatically varies depending on different factors, \textit{e.g.}, 
the LiDAR resolution and range, the scene and the distance. 
Let us assume that each point comes from one of $C$ different groups, representing body parts such as the head, arms, and legs, and the groups are the Gaussian distributions in a mixture (GMM).
Then the set $\lambda$ of parameters of $C$ component GMM is defined as $\lambda=\left\{\left(w_{c}, \mu_{c}, \Sigma_{c}\right), c=1, \ldots C\right\}$, where $w_{c}, \mu_{c}, \Sigma_{c}$ are the weight in mixture, mean, and covariance matrix of $c^{th}$ distribution, respectively.
Different Gaussians are pre-defined and positioned on 3D grids with equal weights and standard deviations.
The likelihood of a single 3D point belonging to the $c^{th}$ Gaussian is:
\begin{equation}
u_{c}(\boldsymbol{p})=\frac{1}{(2 \pi)^{D / 2}\left|\Sigma_{c}\right|^{1 / 2}} \exp \left\{-\frac{1}{2}\left(\boldsymbol{p}-\mu_{c}\right)^{\prime} \Sigma_{c}^{-1}\left(\boldsymbol{p}-\mu_{c}\right)\right\}
\end{equation}
The likelihood of a point belonging to the GMM density is defined as:
\begin{equation}
u_{\lambda}(\boldsymbol{p})=\sum_{c=1}^{C} w_{c} u_{c}(\boldsymbol{p})
\end{equation}
Given a specific GMM, and under the common independence assumption \cite{fisher}, the Fisher vector, $G_{\lambda}^{X}$, can be written as the sum of normalized gradient statistics, computed here for each point $p_t$:
\begin{equation}
G_{\lambda}^{X}=\sum_{t=1}^{T} L_{\lambda} \nabla_{\lambda} \log u_{\lambda}\left(\boldsymbol{p}_{t}\right)
\end{equation}
where $L_{\lambda}$ is the square root of the inverse Fisher Information Matrix \cite{fisher}. We change the variables, from $ w_{c}$ to $\alpha_{c}$, ensuring that $u_{\lambda}(x)$  is a valid distribution and simplifying the gradient calculation:
\begin{equation}
w_{c}=\frac{\exp \left(\alpha_{c}\right)}{\sum_{j=1}^{C} \exp \left(\alpha_{j}\right)}
\end{equation}

Therefore, the normalized gradients can be written as:
\begin{equation}
\mathcal{G}_{\alpha_{c}}^{X}=\frac{1}{\sqrt{w_{c}}} \sum_{t=1}^{T}\left(\gamma_{t}(c)-w_{c}\right)
\end{equation}
\begin{equation}
\mathcal{G}_{\mu_{c}}^{X}=\frac{1}{\sqrt{w_{c}}} \sum_{t=1}^{T} \gamma_{t}(c)\left(\frac{\boldsymbol{p}_{t}-\mu_{c}}{\sigma_{c}}\right)
\end{equation}
\begin{equation}
\mathcal{G}_{\sigma_{c}}^{X}=\frac{1}{\sqrt{2 w_{c}}} \sum_{t=1}^{T} \gamma_{t}(c)\left[\frac{\left(\boldsymbol{p}_{t}-\mu_{c}\right)^{2}}{\sigma_{c}^{2}}-1\right]
\end{equation}
The Fisher vector is formed by concatenating all of these components:
\begin{equation}
\mathcal{G}_{F V_{\lambda}}^{X}=\left(\mathcal{G}_{\alpha_{1}}^{X}, \ldots, \mathcal{G}_{\alpha_{c}}^{X}, \mathcal{G}_{\mu_{1}}^{X^{\prime}}, \ldots, \mathcal{G}_{\mu_{c}}^{X^{\prime}}, \mathcal{G}_{\sigma_{1}}^{X^{\prime}}, \ldots, \mathcal{G}_{\sigma_{c}}^{X^{\prime}}\right)
\end{equation}
To avoid the variation in the number of 3D points in each segment, the resulting FV is normalized by the sample size T:
\begin{equation}
\mathcal{G}_{F V_{\lambda}}^{X} \leftarrow \frac{1}{T} \mathcal{G}_{F V_{\lambda}}^{X}
\end{equation}

Additionally, FV ensures that the extracted features are invariant to input permutation by leveraging symmetric functions.
More specifically, FV calculates the summation of the gradients, which is a symmetric function. 
We extend the basic FV by computing additional symmetric functions including minimum and maximum, as inspired by the max-pooling in~\cite{qi2017pointnet}.
This yields a more descriptive and permutation-invariant representation. 
As a result, each human segment is mapped into a $20\times54$ feature matrix, where $54$ is the number of Gaussians and 20 is the number of features as:
\begin{equation}
F V_{\lambda}^{X}=\left[\begin{array}{l}\left.\sum_{t=1}^{T} L_{\lambda} \nabla_{\lambda} \log u_{\lambda}\left(p_{t}\right)\right|_{\lambda=\alpha, \mu, \sigma} \\ \max _{t}\left(\left.L_{\lambda} \nabla_{\lambda} \log u_{\lambda}\left(p_{t}\right)\right|_{\lambda=\alpha, \mu, \sigma}\right. \\ \left.\min _{t}\left(L_{\lambda} \nabla_{\lambda} \log u_{\lambda}\left(p_{t}\right)\right)\right|_{\lambda=\mu, \sigma}\end{array}\right]
\end{equation}

\subsubsection{Similarity Calculation}

In order to classify the extracted features based on the similarity between human segments, we use deep metric learning to learn a transformation neural network to the embedding space.
As a deep-metric learning method, we use Triplet loss \cite{wang2014learning}. Triplet loss increases the distance between samples of different classes and decreases it between those of the same class.
Cosine similarity is used as the metric function for Triplet loss.

\begin{figure}[t]
    \centering
    \includegraphics[width=\linewidth]{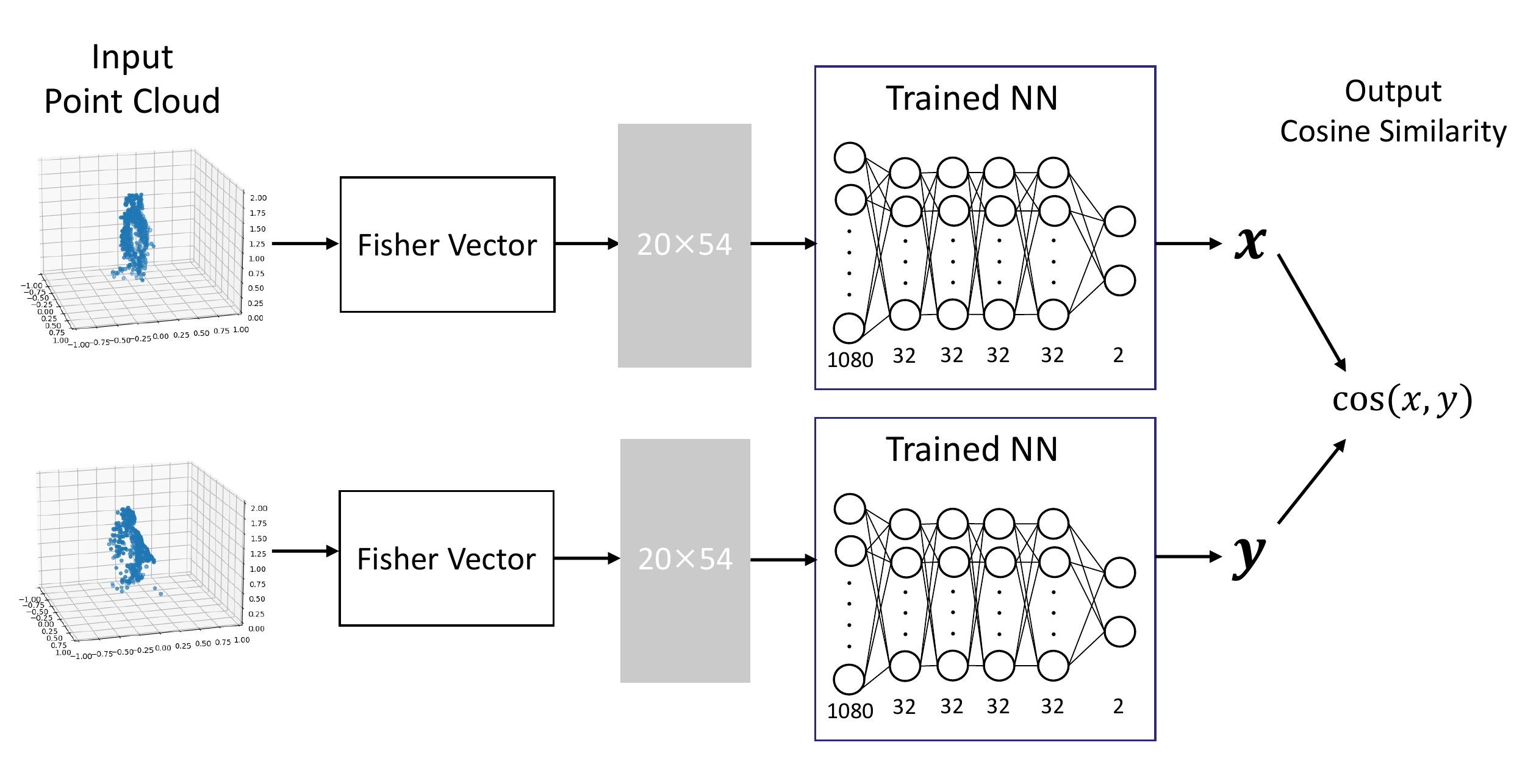}
    \vspace{-0.6cm}
    \caption{Similarity Calculation of Two Human Segments}
    \label{fig:pc_method}
     \vspace{-0.5cm}
\end{figure}

The overall flow is shown in Figure \ref{fig:pc_method}, where we use two human segments (input point clouds) to compute Fisher Vectors. The results are then fed into the trained neural network to obtain the ``coordinates'' in the embedding space. 
The output is the cosine similarity $cos(x,y)$ of the obtained coordinates $x$ and $y$.
Cosine similarity usually ranges in $[-1, 1]$, but we want to use it in the range of [0.1] for consistency with the other features explained later.
Therefore, $P_1$ is defined as follows,
\begin{equation}
    \label{costosim}
    P_1 = \text{similarity}(h_i,h_j) = \frac{\text{cos}(h_i,h_j)+1}{2} 
\end{equation}
where $h_1$ and $h_2$ are human segments in two sub-trajectories of interest, respectively. 

\subsection{Spatial Feature}
\label{subsec:spatial}

The spatial feature of two sub-trajectories represents 
how frequently similar transitions occurred in the past.
For this purpose, we focus on the boundary of each LiDAR $i$'s trajectory area, $\stwo{i}$.

Leveraging the target area map, for each LiDAR $i$, we may find the parts of the boundary of $\stwo{i}$ where pedestrians are likely to enter and leave, which are called \textit{virtual gates}.
A virtual gate may be a door, the boundary of $\stwo{i}$ on hallways, 
and so on, 
and for each sub-trajectory, its start point (or end point) should belong to a virtual gate. 
Then we build a transition matrix $\trans$ where each element is a transition probability from one virtual gate to another.

Given a destination virtual gate, say $g_2$, 
to which the starting point of $tr_v$ belongs, 
we define the spatial feature probability $P_2$ of sub-trajectories $tr_u$ and $tr_v$ as the ratio of 
the visit from the virtual gate, say $g_1$, to which the end point of $tr_u$ belongs over the sum of all the probabilities from the other gates to $g_2$. 
This is defined as follows: 
\begin{equation}
    \label{transition_prob}
    P_2 =\frac{\trans(g_{1},g_{2})}{\sum_{k}\trans(g_{k},g_{2})}
\end{equation}

\subsection{Temporal Feature}
\label{subsec:temporal}

Similarly, the temporal feature of two sub-trajectories is defined 
to represent the likelihood of the traveling time from one virtual gate to another. 
Let $\traj{[t_a, t_b]}{i}{u}$ and $\traj{[t_c, t_d]}{j}{v}$ 
denote the two sub-trajectories. 
The traveling time is obtained as $\Delta t = t_c - t_b$ if $\traj{[t_a, t_b]}{i}{u} < \traj{[t_c, t_d]}{j}{v}$. 
Assuming prior probability density function $\timedist{x}$, 
we can obtain the probability by 
\begin{equation}
    \label{eq:temporal}
    P_3 = \timedist{\Delta t}
\end{equation}

\subsection{Spatial and Temporal Feature Distributions Update}
\label{subsec:beyesian}

Finally, we describe how to update the transition matrix $\trans$ (spatial feature) and 
probability density function $\timedist{x}$ (temporal feature). 
The former is done by a histogram, and the latter is based on the Bayesian system. 

These functions should be updated with high confidence during the operation, and one good phenomenon is to believe the case 
with only one pedestrian traveling from one end point to another starting point and no other pedestrian is observed. 
This phenomenon may happen in less crowded scenes (\textit{e.g.}, early morning). 
The transition matrix can easily be updated by the recorded histogram of the past transitions with high confidence. 
For Bayesian updating of travel time probability distribution, the likelihood of travel time $P(E|H)$ in such a case with high confidence is similar to a normal distribution.
Therefore, the prior distribution is updated using a Bayesian formula shown in Formula (\ref{baysian}).
\begin{equation}
    \label{baysian}
    P(H|E)=\frac{P(E|H)\cdot P(H)}{P(E)}
\end{equation}
Here, the prior distribution of travel time $P(H)$ and the posterior distribution $P(H|E)$ created are both calculated as inverse gamma distributions.

% ===
% Evaluation
% ===
\section{Evaluation}
\label{sec:evaluation}

In this section, we evaluate the proposed system 
using our testbed. We have installed 70 LiDARs 
in our university's new campus building, covering from the 1st floor (= ground floor) to the 6th floor. It took almost two years for design, implementation, and installation, and we have just started collecting human trajectory data.

\subsection{System Specification, Environment and Dataset}

The specifications of LiDARs are summarized in Table 1.

\begin{table}[tb]
  \caption{3D LiDAR specifications}
    \vspace{-0.25cm}
  \centering
  \begin{tabular}{l|l|l} \hline \hline
     & Livox Avia & Hokuyo \\ 
     &&YVT-35LX\\\hline 
    Maximum number of points (point/frame) & 240,000  & 2,664 \\
    Frame rates (frame/s) & 10  & 10 \\
    Maximum detection distance (m) & 460 &  35 \\
    Horizontal field of view angle (\textdegree) & 70.4  & 210 \\
    Vertical field of view angle (\textdegree) & 77.2  & 40 \\
    Distance precision (1$\sigma$ at 20m)  (cm) & $\pm$ 2.0 & $\pm$ 0.1 \\
    Angular precision  (1$\sigma$) (\textdegree) & $\pm$ 0.05 & $\pm$ 2.0 \\ \hline
  \end{tabular}
  \label{tab:livox}
   \vspace{-0.5cm}
\end{table}

In this paper, we conducted the following experiments at three on the campus to evaluate (i) basic performance under intended controlled scenarios (at the mid-size indoor square on 2F, \textbf{Experiment-1}), 
(ii) in-situ performance evaluation with high-density, narrow FoV LiDARs (similar to RGB camera) at the mid-size entrance hall on 1F for comparison with the RGB camera-based method, \textbf{Experiment-2}), 
and 
(iii) in-situ performance evaluation with wide FoV and relatively low-density LiDAR in 
a long corridor with a lot of lecture rooms (5F, \textbf{Experiment-3}). 
These experimental environments are shown in Fig. \ref{fig:maps} and \ref{fig:photos}.

Comprehensive investigations in Experiment-1 include the accuracy variation according to congestion (number of persons) (\textbf{Scenario 1-(a)}),  performance improvements with the proposed update method (\textbf{Scenario 1-(b)}),  component-by-component performance measurements (\textbf{Scenario 1-(c)}).

\begin{table}[tb]
  \caption{Dataset Statistics\color{black}}
  \centering
    \vspace{-0.25cm}
  \begin{tabular}{l|c|c|c} \hline \hline
      & Experiment-1 & Experiment-2 & Experiment-3 \\ \hline
     \# of observed persons & 32 (max) & 2,356 & 15,101 \\
     \# of sub-trajectories & 319 (max) &  4,062  & 19,356\\
     \# of switches         & 287 (max)  &  1,706  & 4,255 \\ 
     \hline
  \end{tabular}
  \label{tab:dataset}
   \vspace{-0.5cm}
\end{table}

\subsection{Data Collection}
The statistics of dataset obtained is described in Table. \ref{tab:dataset}
\color{black}

\begin{figure*}[t]
\begin{center}
\subfigure[2nd Floor (Experiment-1)]{%
    \label{fig:2fmap}
	\includegraphics[width=0.56\columnwidth]{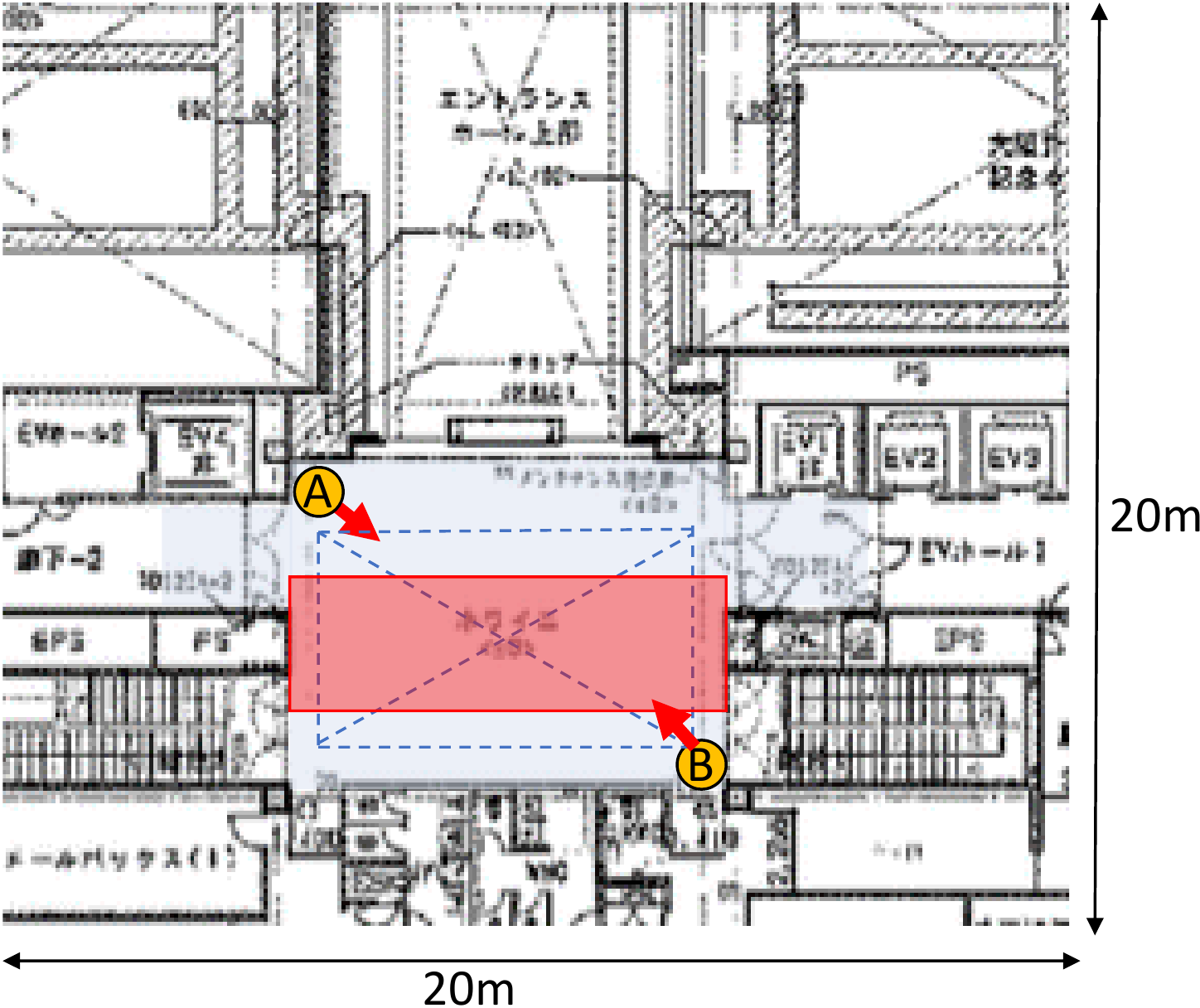}}%
	\hspace{1.5cm}
\subfigure[1st Floor (Experiment-2)]{%
    \label{fig:1fmap}
	\includegraphics[width=0.5\columnwidth]{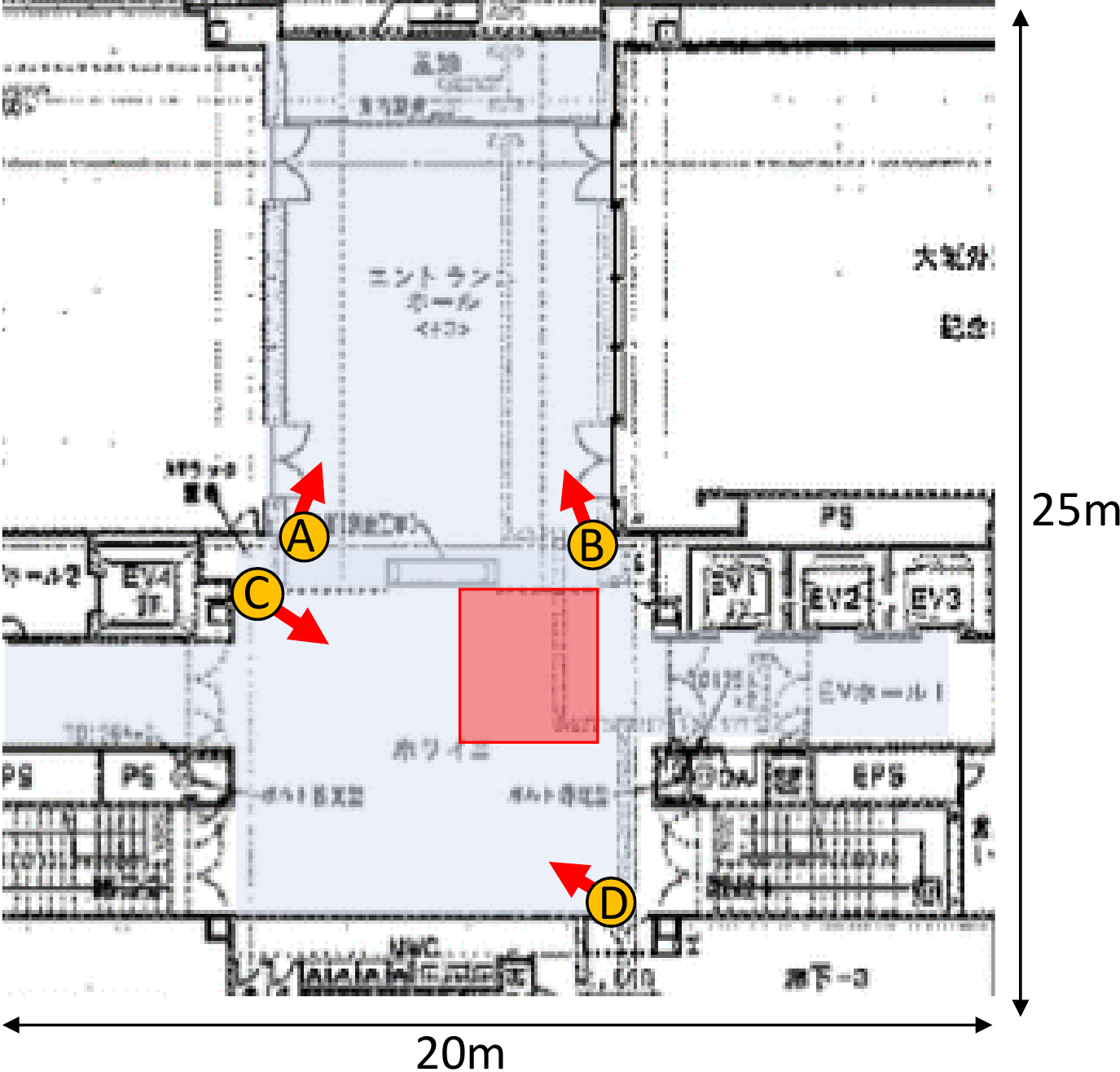}}%
	\hspace{1.5cm}
\subfigure[5th Floor (Experiment-3)]{%
    \label{fig:5fmap}
	\includegraphics[width=0.5\columnwidth]{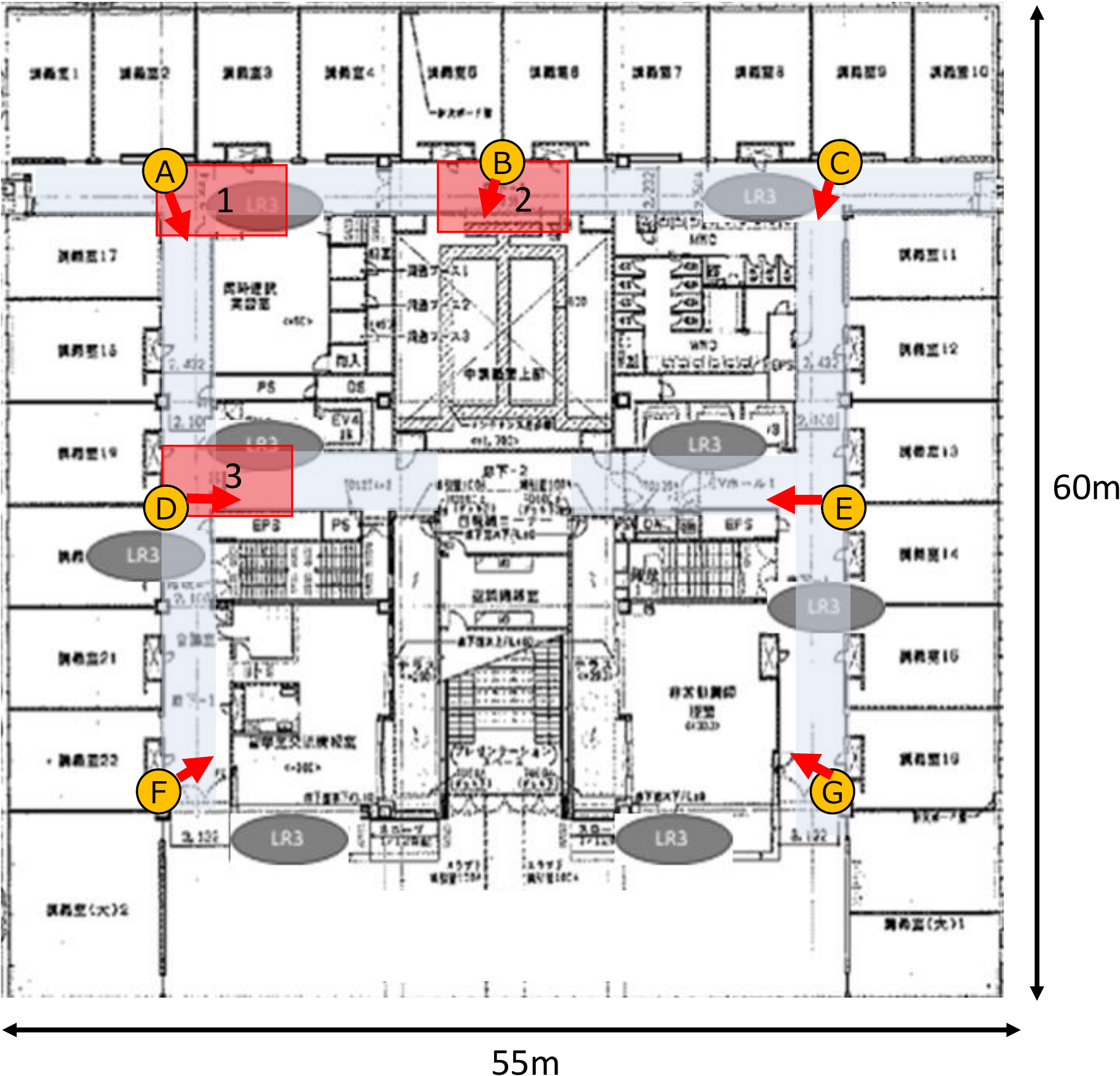}}%	
\centering

\caption{Maps of the experimental environments. Yellow circles and red arrows indicate LiDAR locations and directions, respectively. The total visible range of LiDARs' are filled with light blue, and the eliminated regions for evaluations are illustrated as red rectangles. Each side of the red rectangle corresponds to the ``virtual gates''.  (b) Blue dotted lines represent the real trajectory.}
\label{fig:maps}
\end{center}
\vspace{-0.5cm}
\end{figure*}

\begin{figure*}[t]
\begin{center}
\subfigure[2nd Floor (Experiment-1)]{%
    \label{fig:2fphoto}
	\includegraphics[width=0.57\columnwidth]{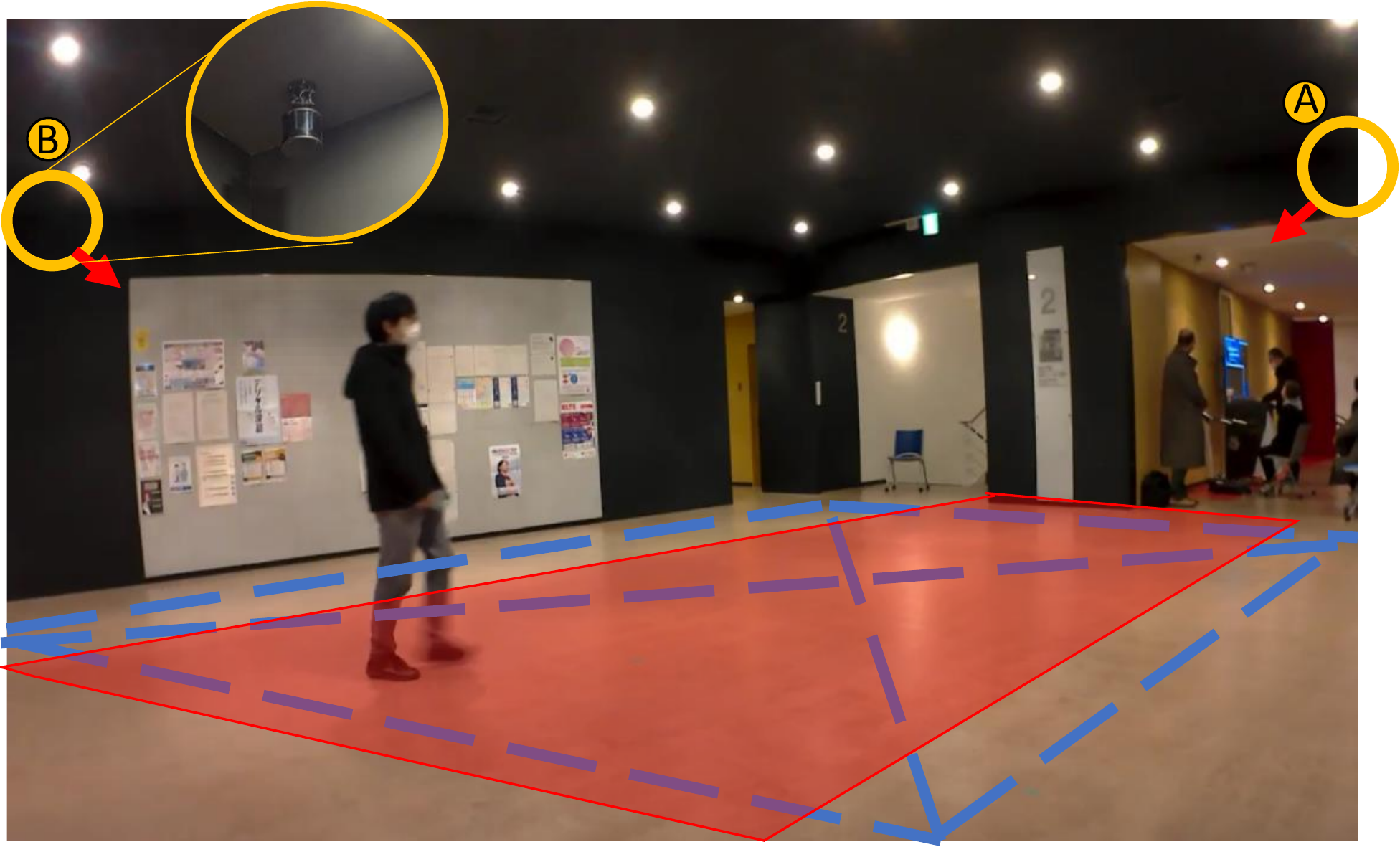}}%
	\hspace{1.0cm}
\subfigure[1st Floor (Experiment-2)]{%
    \label{fig:1fphoto}
	\includegraphics[width=0.5\columnwidth]{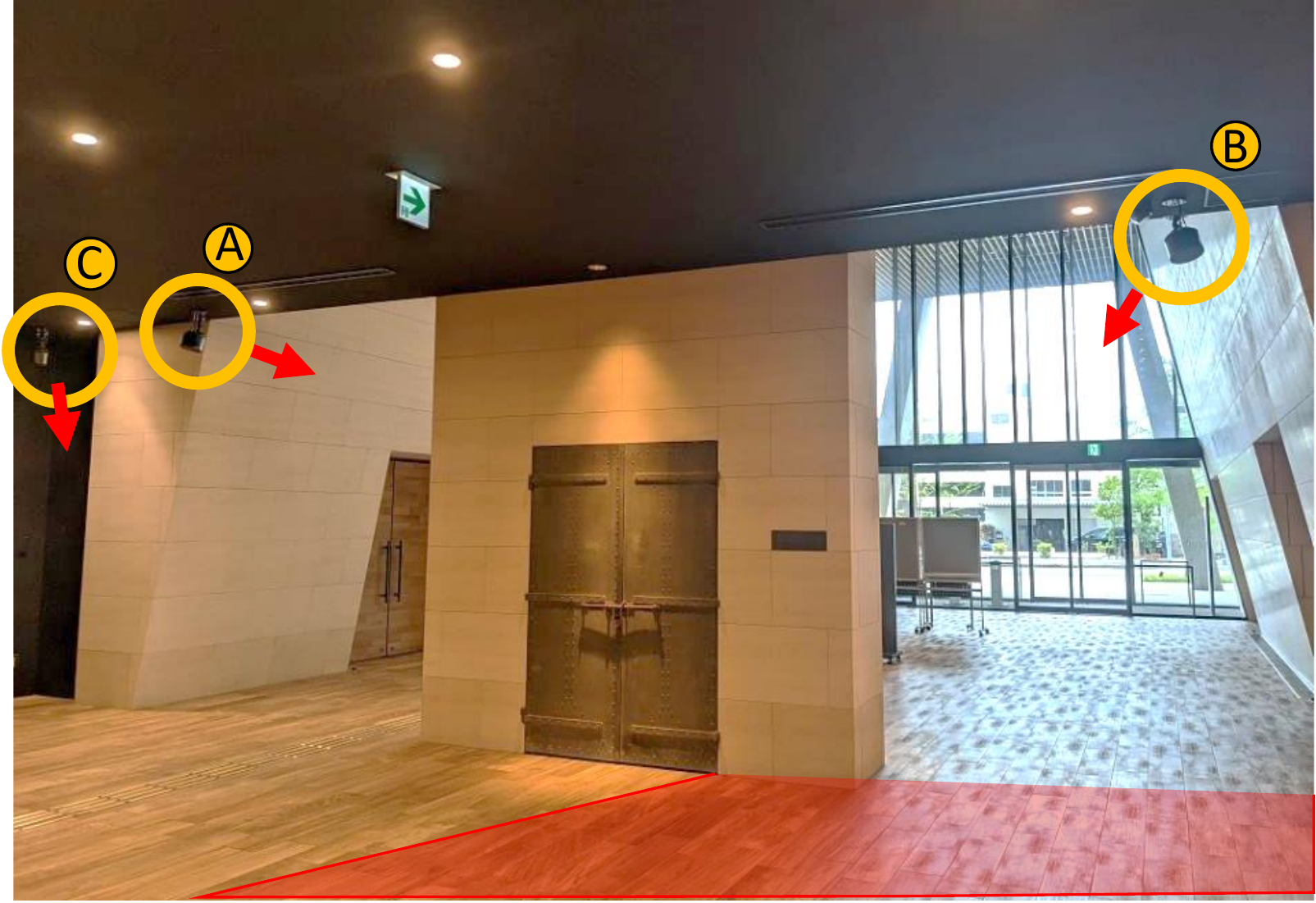}}%
	\hspace{1.0cm}
\subfigure[5th Floor (Experiment-3)]{%
    \label{fig:5fphoto}
	\includegraphics[width=0.54\columnwidth]{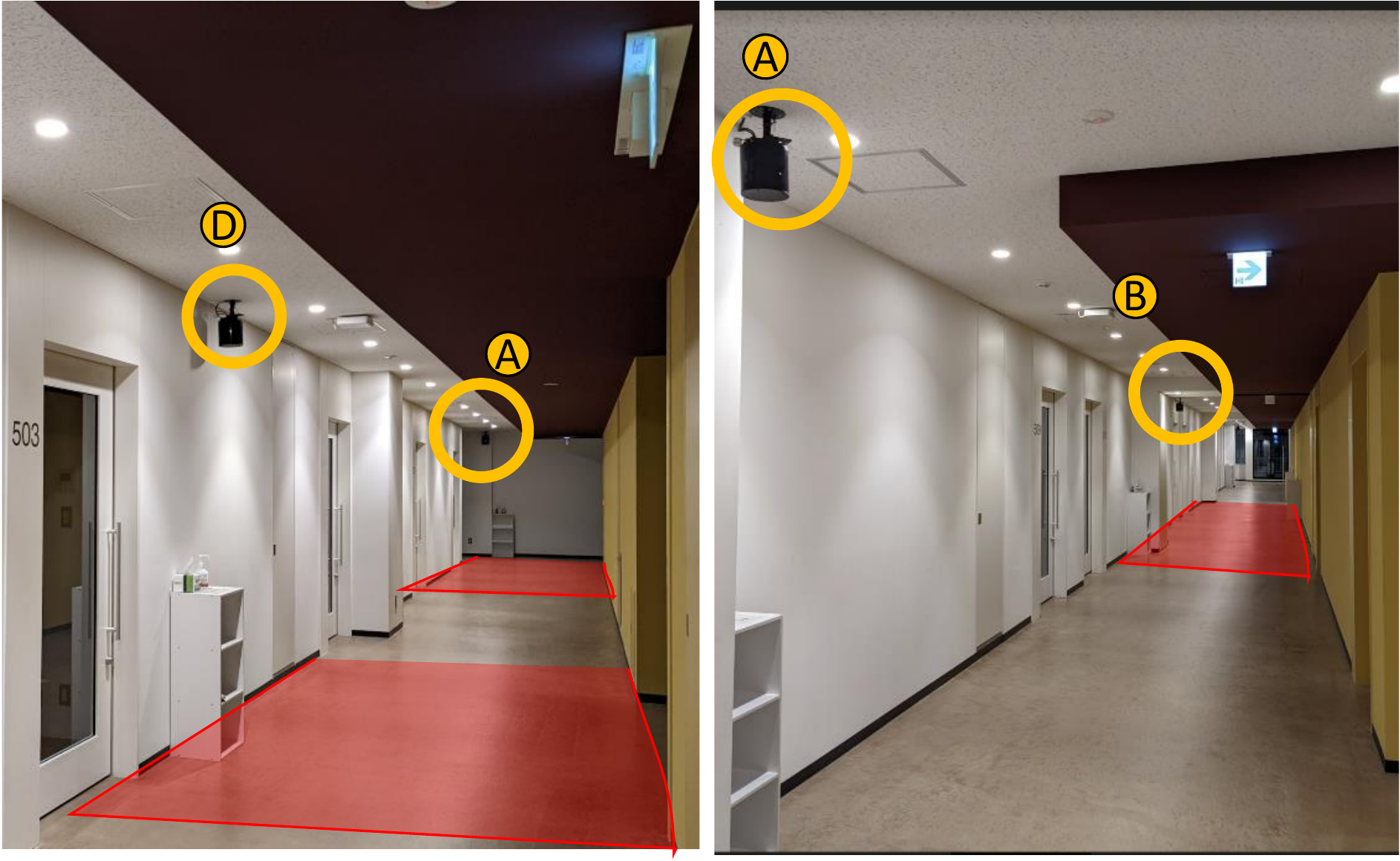}}%	
\centering
\caption{Experimental environments. Lines and regions correspond to the maps.}
\label{fig:photos}
\vspace{-0.5cm}
\end{center}
\end{figure*}

In Experiment-1, we recruited 32 general subjects with different distributions of genders and ages (20's--50's).
Each subject was asked to walk on a designated route among the four turning points. The rectangle by those points is $4m\times 7m$ as shown in Fig. \ref{fig:2fmap}, \ref{fig:2fphoto}.
For each subject, we collected approximately 1000 frames of 3D point clouds.
In this experiment, the LiDAR beams are not occluded by other subjects.
This means that the sub-trajectories are clearly obtained, and we can evaluate the pure matching performance with complete sub-trajectories.  

In Experiment-2, we observed residents and visitors of the building at the entrance hall using 4 LiDARs (Livox Avia).
The trajectories, eliminated space (the red rectangle) and trajectory areas are shown in Fig. \ref{fig:1fmap}, \ref{fig:1fphoto}.
In total, 2,356 complete trajectories (\textit{i.e.}2,356 pedestrians)  were observed from September 10 to September 13, 2022 (4 days). 
Most importantly, the comparison with the RGB camera-based method was conducted in Experiment-2. 

In Experiment-3, similar to Experiment-2,  
we observed the residents and visitors of the building on the 5th floor with one corridor with a lot of rooms (Fig. \ref{fig:5fmap}, \ref{fig:5fphoto}).
However, the different LiDARs (Hokuyo YVT-35LX) are installed on 5F as we need a more horizontal view (wide range). We have used 7 Hokuyo LiDARs to track pedestrians. The key signatures of this experiment are 
different mobility (longer trajectories), different LiDARs (lower point cloud density compared with the former experiments), and the wider area.

Although almost the entire floor in each experiment is captured by the installed LiDARs, we intentionally eliminate the point cloud in the center area represented by the red rectangle in Fig.  \ref{fig:maps}, and evaluated our method using these area as LiDAR-blank regions.
\color{black}
More detailed configurations are explained in the following subsections.

\subsection{Experiment-1: Scenarios and Results}

The recruited 32 subjects walked independently, following the same route.
Then we synthesized the point clouds of multiple subjects to generate multiple scenarios with the different numbers of subjects with different timings. 

\subsubsection{Scenarios}

In \textbf{Scenario 1-(a)}, 
we synthesized, we generated different numbers (2, 4, 8, 16, and 32) of subjects, where we delayed for 10 seconds the start time of the following subjects to make intervals between subsequent subjects. 
In \textbf{Scenario 1-(b)}, 
we changed the delay time (0, 5, 10, 15, and 20 seconds). 
This scenario mainly aims to assess the effect of travel time distribution. That is, the shorter the delay time is, the harder it is to distinguish travel time.
In \textbf{Scenario 1-(c)}, to evaluate the contributions of each features
($P_1$, $P_2$ and $P_3$), 
the matching is performed only with one of the three features. 

In all the scenarios, the matching performance is compared before and after the spatial and temporal features (transition matrix $\trans$ and travel time probability distribution $\timedist{x}$) is updated.
As their initial values, all the probabilities in $\trans$ and $\timedist{x}$ are uniform (we assumed a certain range for $\timedist(x)$)
and the observations obtained through all the scenarios are used to update the both.

\subsubsection{Result in Scenario 1-(a)}

We show F-measure for each number of subjects in Figure \ref{fig:f-difpeople}, where before and after the updates of the transition matrix and 
travel time distribution are shown. 
As the initial distribution is fully uniform (zero knowledge), the accuracy with 32 subjects (this is an extreme (highly-crowded) case where 32 people in $28m^2$ \cite{NIKOLIC201658})) is around 0.6, but after the update, it becomes much better. 
With the normal walking speed, 4-8 subjects generate appropriate densities. By looking at the values, the F-measure after the update is 0.8-0.9, which is sufficiently high.
Based on the observation above, we chose the four-subject case for Scenario 1-(b).

\subsubsection{Result in Scenario 1-(b)}

\begin{figure}[t]
\centering
\subfigure[in Scenario 1-(a) (with Different Number of Subjects, Interval=10sec.)d]{
    \includegraphics[clip, width=0.46\linewidth]{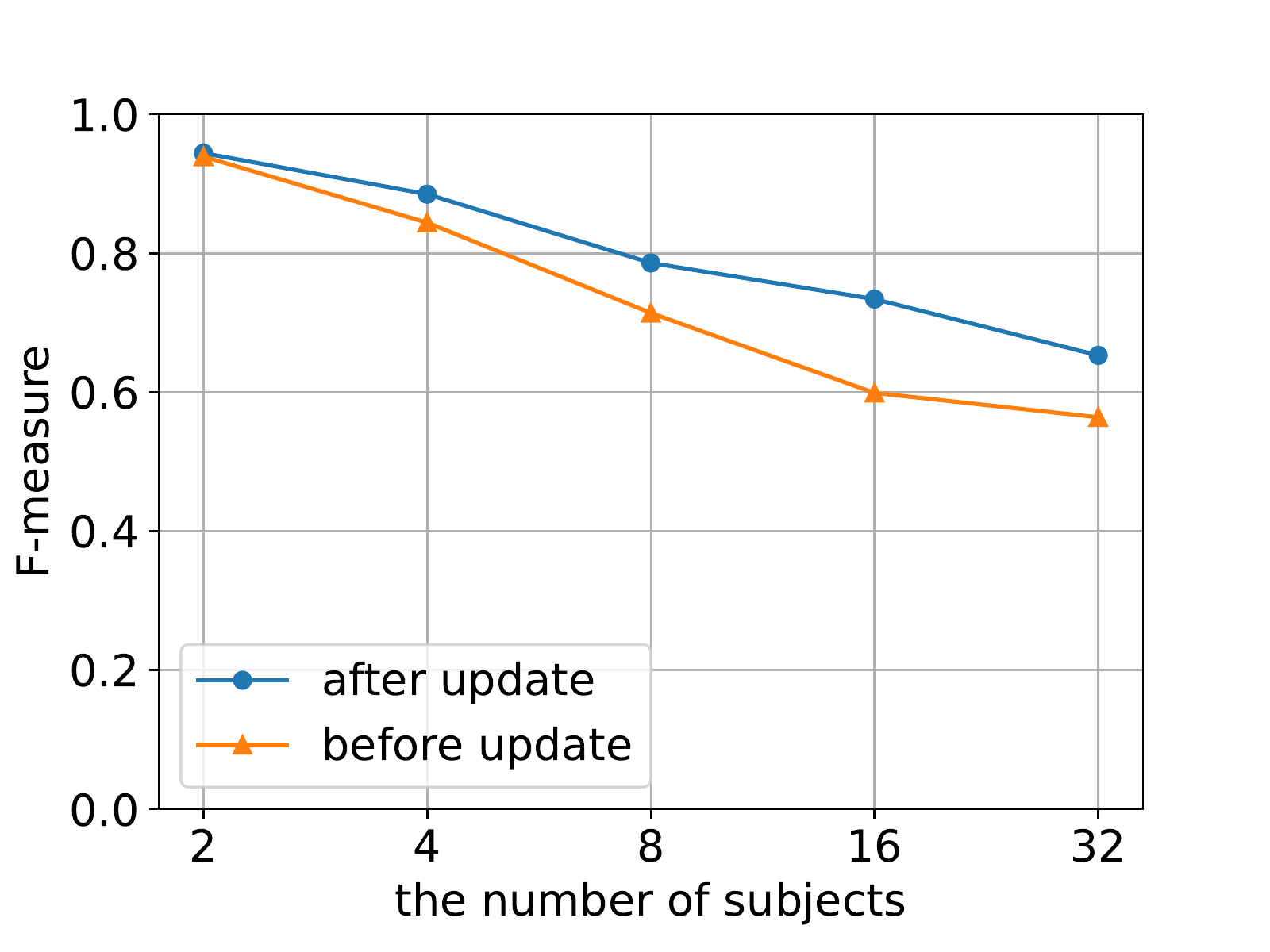}
    \label{fig:f-difpeople}
}
\subfigure[in Scenario 1-(b) (with Different Time Intervals, \# Subjects = 4)]{
    \includegraphics[clip, width=0.46\linewidth]{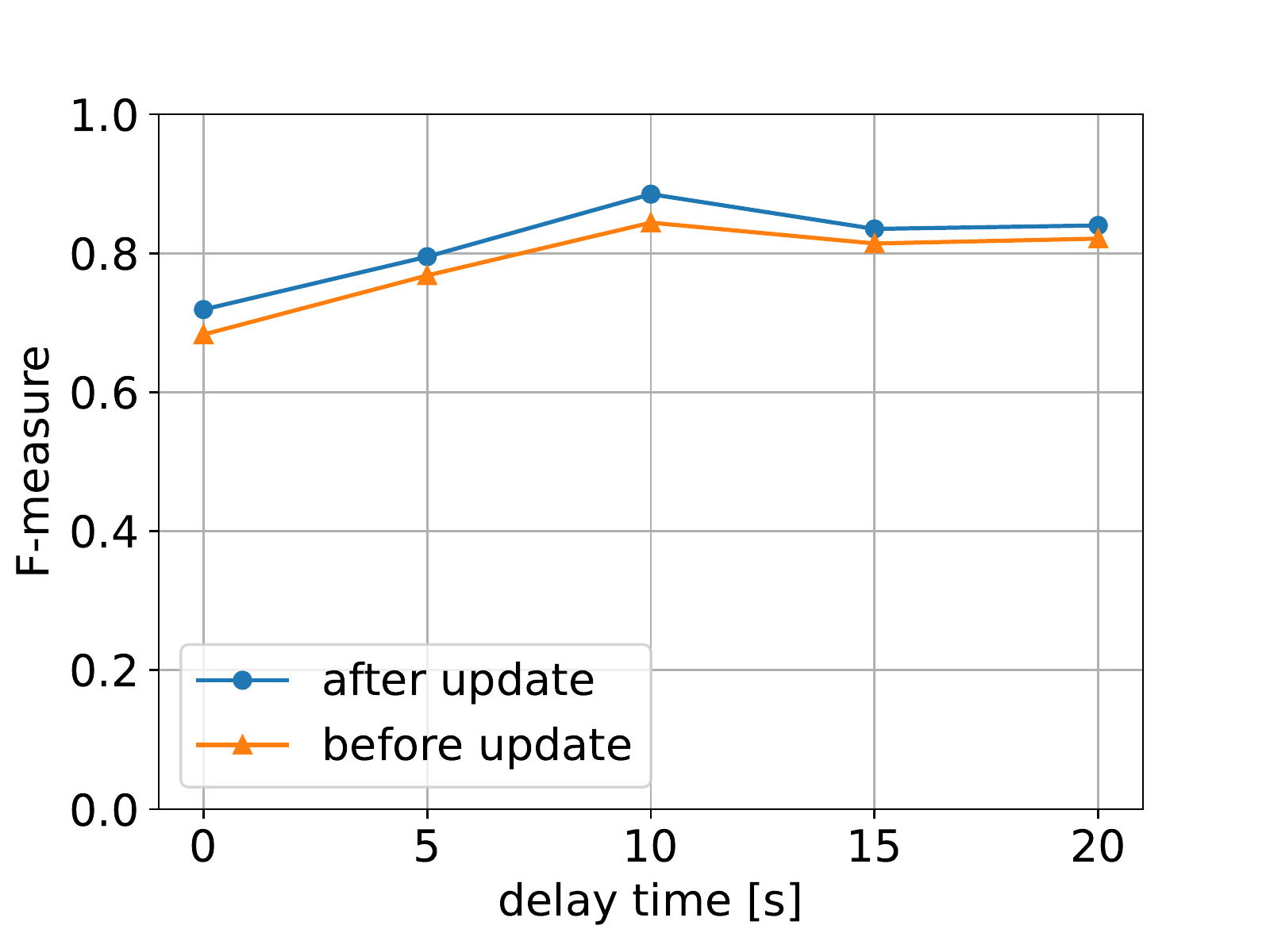}
    \label{fig:f-diftime}
}
\caption{F-measure variation}
\end{figure}

We also show F-measure for each interval time in Fig. \ref{fig:f-diftime}. 
With a longer interval, we achieved higher accuracy till 10 sec., which is very natural. 
We also see a decrease with longer intervals, and this means that a 10-second interval was optimal for making distinguishable features 
in travel time and patterns. 
This entirely depends on the scenarios.

\subsubsection{Result in Scenario 1-(c)}
We evaluated F-measure of the cases with only one feature of $P_1$, $P_2$ and $P_3$, and the result is shown in Table \ref{tab:f_features}. 
In all the one-feature cases, 
F-measure values are between 0.7 to 0.8, and with the three features, it is 0.89, which showed the effectiveness of the combination of the features.

\begin{table}[t]
    \centering
    \caption{Contributions of Features to Accuracy (\# Subjects=4 and  Interval=10sec.)}
    \vspace{-0.3cm}
    \label{tab:f_features}
    \begin{tabular}{c|c}
        \hline \hline
        Feature&F-measure (Post-update)\\
        \hline
        Point Cloud Feature ($P_1$) &0.74\\
        Spatial Feature  ($P_2$) &0.71\\
        Temporal Feature ($P_3$) &0.80\\
        Combination ($P_1 \cdot P_2 \cdot P_3$) & 0.89\\
        \hline
    \end{tabular}
     \vspace{-0.3cm}
\end{table}

\subsubsection{Effect of Distributions Update}

\begin{table}[t]
    \caption{Accuracy (Pre- and Post-Updates) (\# Subjects=4 and  Interval=10sec.)}
        \vspace{-0.3cm}
    \label{tab:standard_res}
    \centering
    \begin{tabular}{c|c|c|c}
    \hline \hline
         &Precision&Recall&F-measure \\
         \hline
        Pre-update&0.85&0.84&0.84\\
        Post-update&0.89&0.88&0.89\\
        \hline
    \end{tabular}
     \vspace{-0.3cm}
\end{table}

\begin{figure}[t]
\centering
\subfigure[Pre-update]{
    \includegraphics[clip, width=0.46\linewidth]{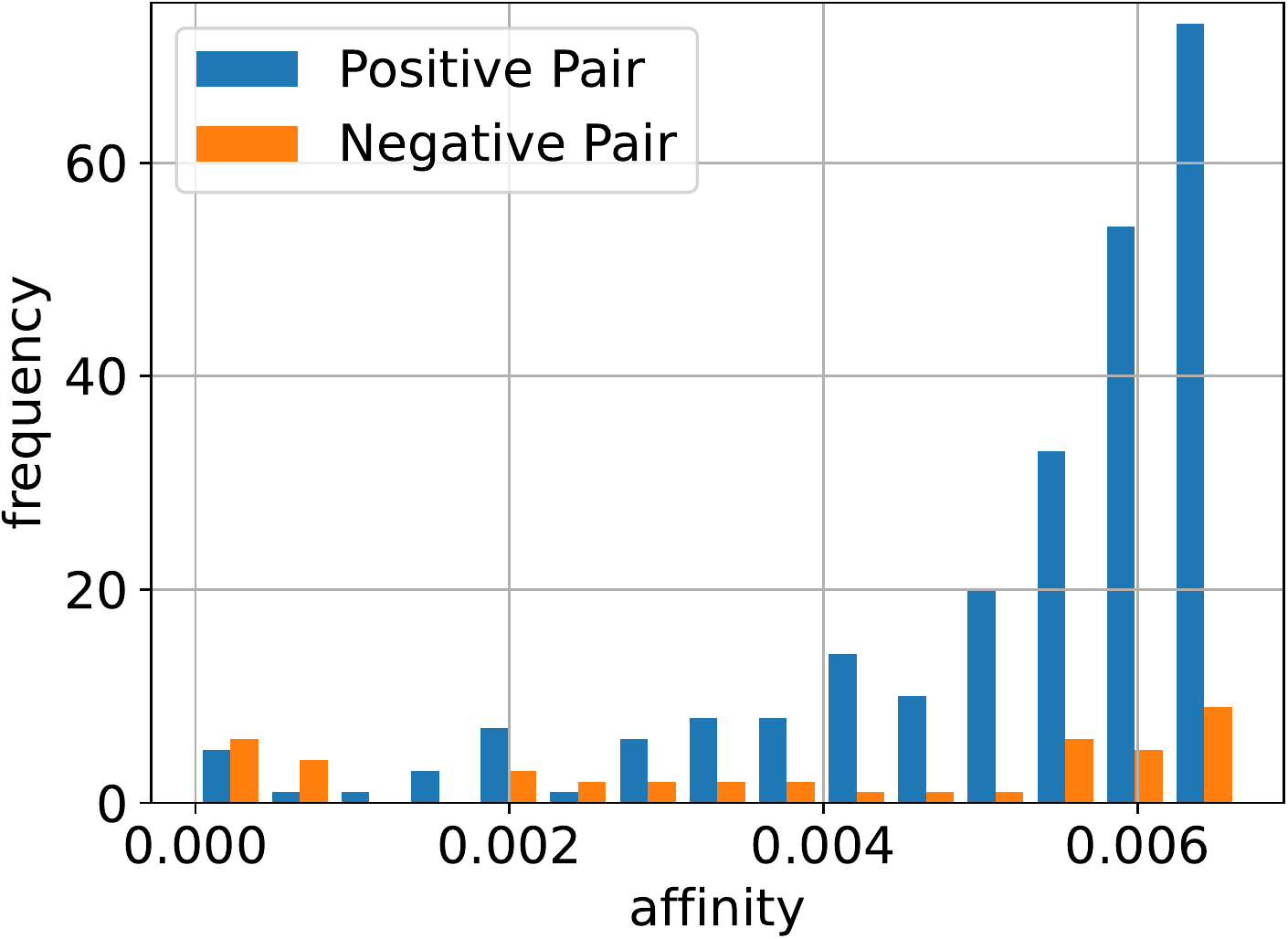}
    \label{fig:pre-affinity}
}
\subfigure[Post-update]{
    \includegraphics[clip, width=0.46\linewidth]{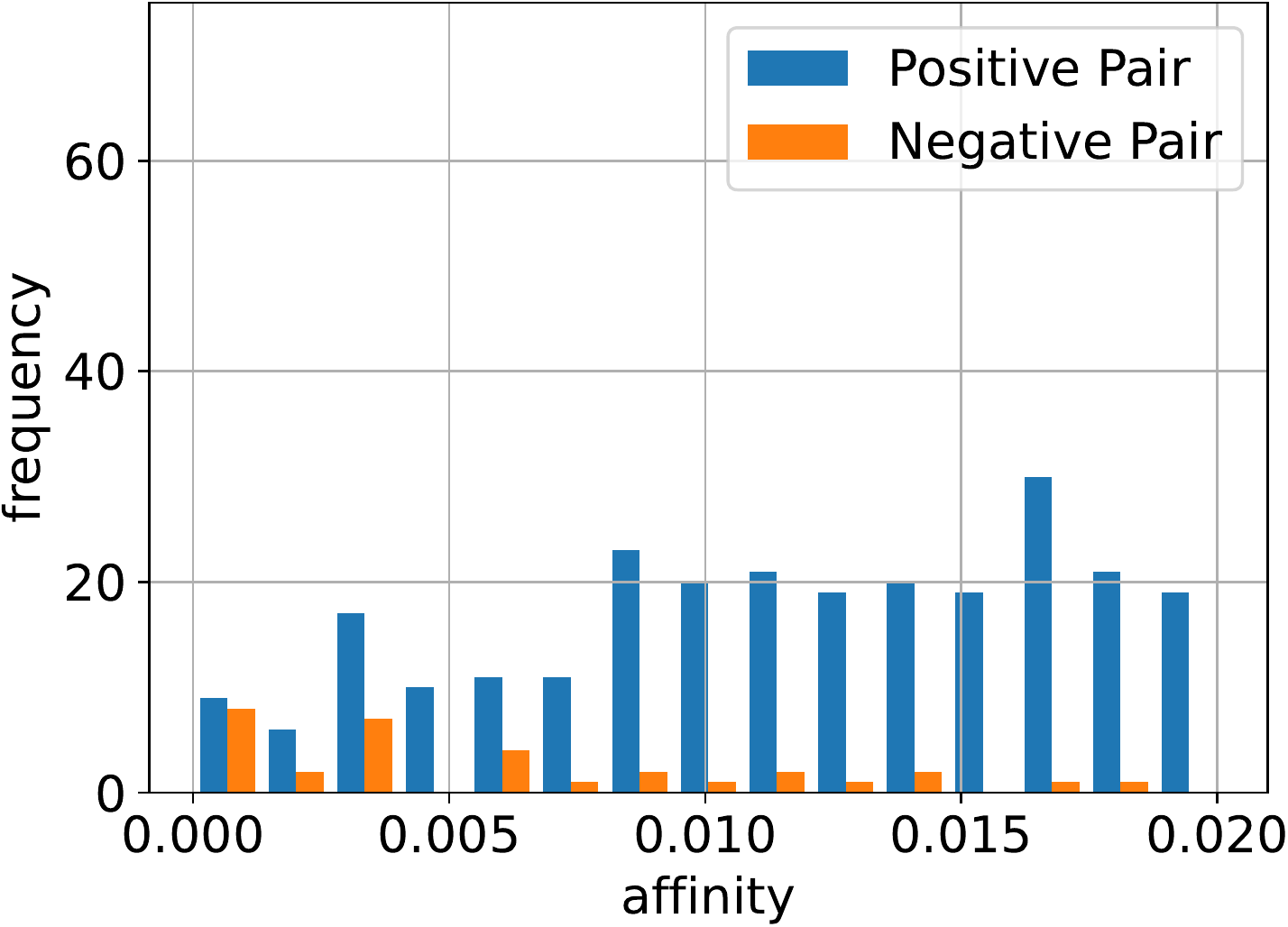}
    \label{fig:post-affinity}
}
\caption{Affinity Distribution}
\vspace{-0.5cm}

\label{fig:affinity_dist}
\end{figure}

\begin{figure}[t]
\centering
\subfigure[Affinity Distribution of Reliable Data]{
    \includegraphics[clip, width=0.46\linewidth]{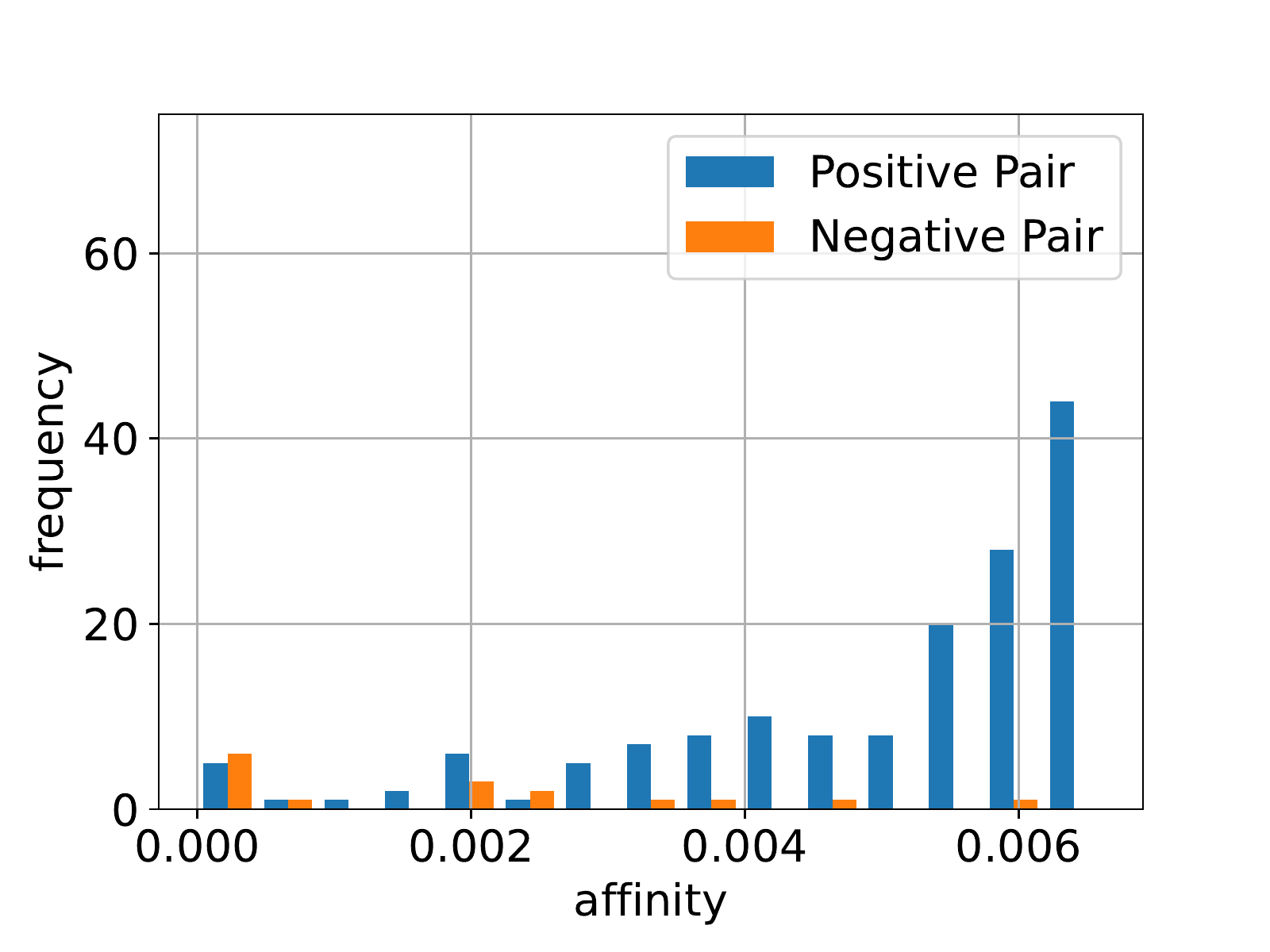}
    \label{fig:reliable}
}
\subfigure[Travel Time Distribution Update]{
    \includegraphics[clip, width=0.46\linewidth]{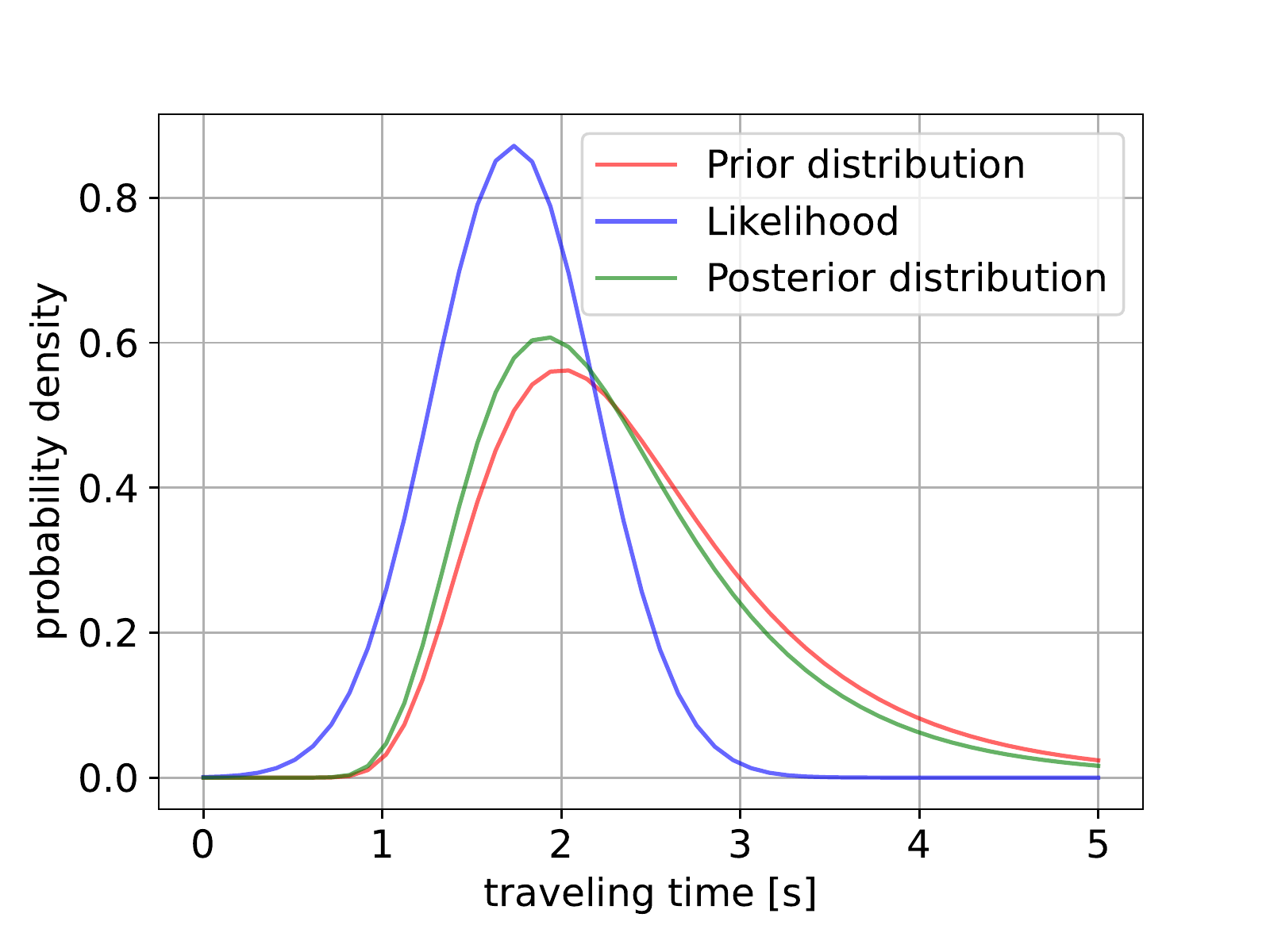}
    \label{fig:travel-time}
}
\vspace{-0.2cm}
{\caption{High Confidence Data for Update}}
\label{fig:reliable-time}
\vspace{-0.3cm}
\end{figure}

Table \ref{tab:standard_res} shows the accuracy before and after updates. 
The updated F-measure has improved to 0.89.
The distribution of affinities for the matched pairs is shown in Fig.  \ref{fig:affinity_dist}. Before the update, negative pairs exist 
in a wide range, converging to lower affinity cases after the update.
The distribution of affinities for the pairs with high confidence is shown in Fig. \ref{fig:reliable}.
We can clearly see high affinities have few negative pairs.
Finally, Fig. \ref{fig:travel-time} shows 
how the distribution is updated in traveling from the right top corner to the left top corner in Fig. \ref{fig:2fmap}.
The prior distribution is an orange curve with high variance and the likelihood is the purple curve with the distribution calculated from the travel with high confidence.
The posterior distribution is the blue curve, which is the result of the Bayesian update.

\subsubsection{Deep Metric Learning Performance}

\begin{figure}[t]
\centering
\subfigure[Similarity Matrix of Subject Pairs (red (close to 1) and blue (close to -1) mean more and less similarity values, respectively)]{
    \includegraphics[clip, width=0.5\linewidth]{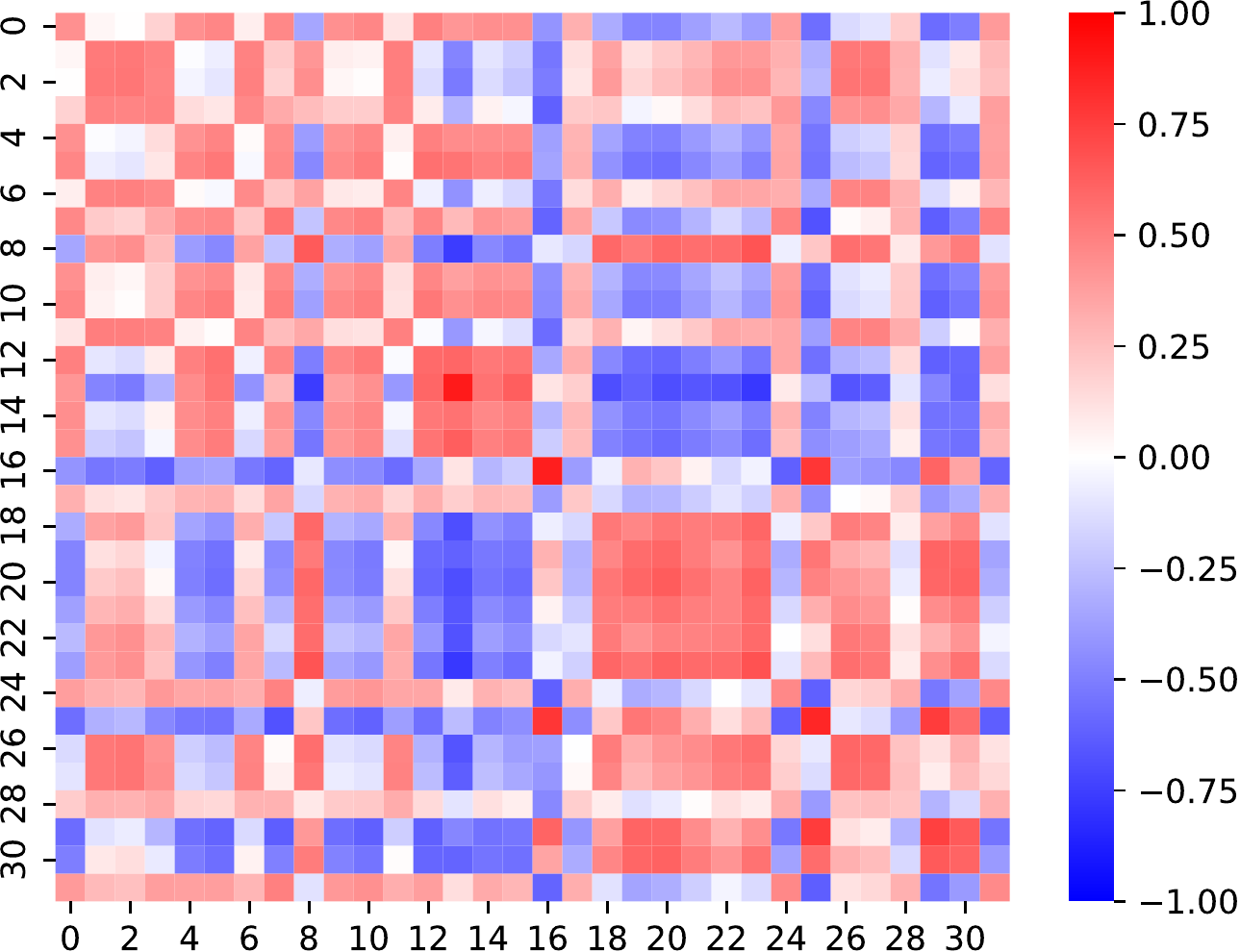}
    \label{fig:heatmap}
}
\subfigure[ROC Curve of Re-id]{
    \includegraphics[clip, width=0.42\linewidth]{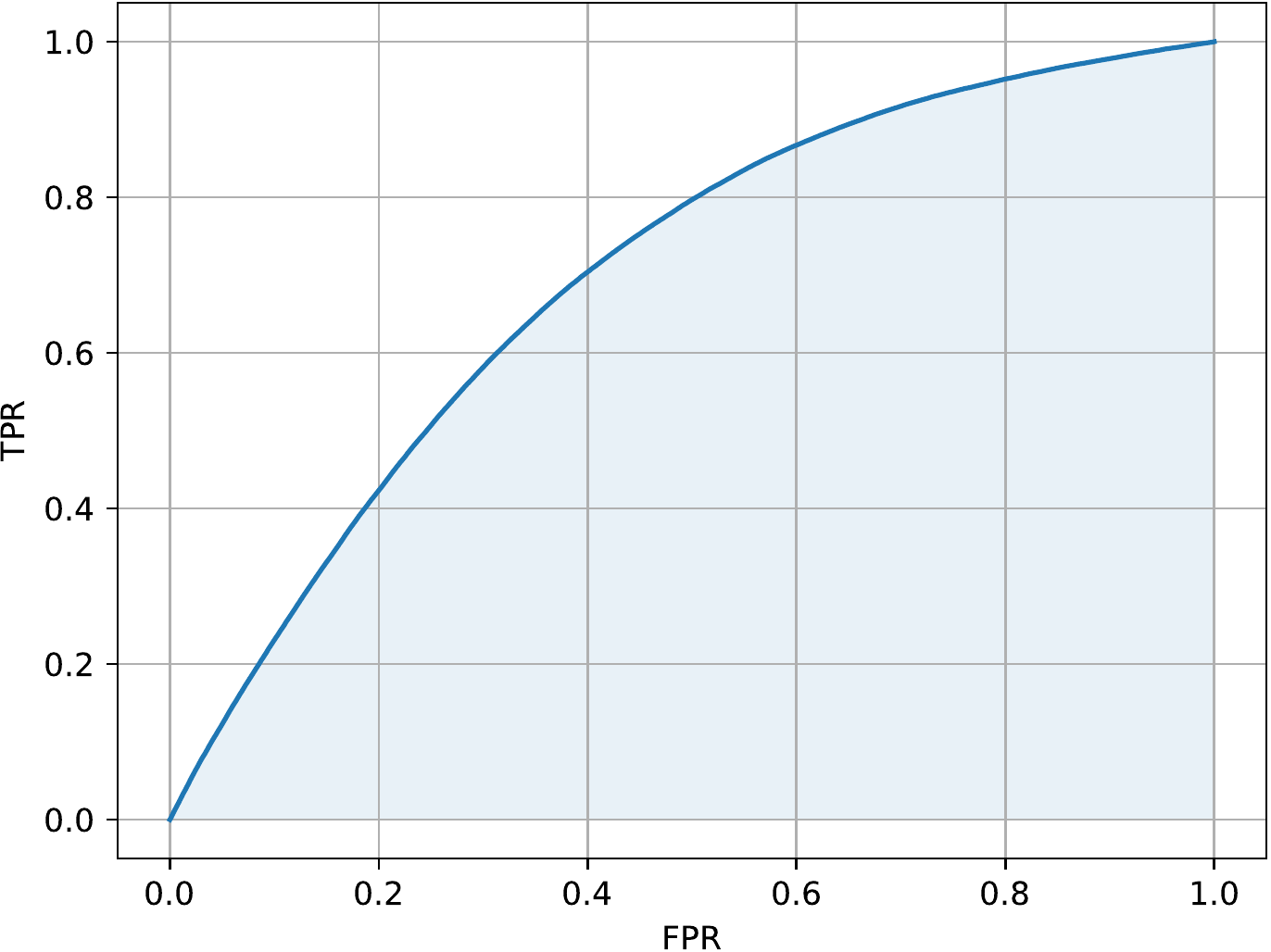}
    \label{fig:roc}
}
{\caption{Performance of deep metric learning}}
\vspace*{-0.5cm}
\label{fig:eval-exp1-DML}
\end{figure}

We investigated the basic performance of deep metric learning over 32 subjects. 
We trained the model on 90\% of the randomly selected data and tested it on the remaining 10\%.
The average cosine similarity value of two human segments contained in the test data was calculated, and the result is shown as the matrix in Fig. \ref{fig:heatmap} where red cells mean higher similarity values.
We can see red cells are seen along the diagonal line from top-left to bottom-right (this means similarity tread is correct), 
but there are also red cells in different cells. 
To quantify the result, the means and standard deviations of the similarity between the same and different persons on the diagonal are shown in Table \ref{tab:sim_dist}. 
From the result, we can clearly see the larger deviation with 
different subjects. 
The ROC curve is also shown in Fig. \ref{fig:roc} where 
AUC (Area Under the Curve) was 0.71. Generally, AUC above 0.7 means pretty high accuracy. 

\begin{table}[t]
    \centering
    \caption{Mean and Standard Deviation of Similarity Values}
    \vspace*{-0.3cm}
    \label{tab:sim_dist}
    \begin{tabular}{c|c|c}
    \hline \hline
         & mean & standard deviation \\
         \hline
        similarity between same person & 0.57 & 0.14\\
        similarity between different persons & 0.02 & 0.41\\
         \hline
    \end{tabular}
     \vspace{-0.5cm}
\end{table}

\subsection{Experiment-2: Comparison with RGB Camera}

In Experiment-2, which involved four days of tracking pedestrians in in-situ environment, the accuracy with the proposed method was fairly high (F-measure=0.98) due to fewer crowds as shown in Table. \ref{tab:dataset}.

We compared our method with a well-known person re-identification system using RGB cameras \cite{zheng2016person}, using this dataset since the tracking situation in Experiment-2 is similar to when using distributed RGB cameras.
LiDARs on the 1st floor are installed in the same position and orientation as the surveillance camera, taking into account the design of the building. Also, the installed LiDAR (Livox Avia) has a relatively similar FoV as the camera.
\color{black}
Since it is not possible to implement the same system, 
we have used the re-id function in \cite{zheng2016person}, 
which is much more accurate, instead of our point cloud-based re-identification. 
On the other hand, if we use the camera system, it is usually possible to obtain the transition pattern as the tracking in a single camera is less accurate than single LiDAR-based tracking. 
Consequently, our method can leverage point cloud features ($P_1$), transition ($P_2$) and travel time ($P_3$), 
while the camera-based method can use color-based features ($P^{+}_1$) 
and travel time ($P_3$). 
The result is shown in Table \ref{tab:cmp}.
The proposed method achieved sufficient accuracy with the help of 
reasonable $P_1$ and original $P_2$, while RGB can achieve higher 
with the power of image-based features. 
Our method achieved a very good trade-off between privacy and accuracy compared with the well-known camera-based re-id method. 

\begin{table*}[t]
\vspace{-0.5cm}
\caption{Comparison with Other Re-identification Methods}
    \label{tab:cmp}
    \centering
    \begin{tabular}{c|c|c|c}
    \hline \hline
        Method & Features to use & F-measure (Pre-update) & F-measure (Post-update) \\
         \hline
        Ours & Point-cloud Re-ID ($P_1$) + Accurate Transition ($P_2$) + Travel Time ($P_3$) & 0.844 & 0.885\\
        RGB \cite{zheng2016person} & RGB-based Re-ID (accuracy++ / privacy{-}{-}) + Travel Time ($P_3$) & 0.851&0.943\\
        \hline
    \end{tabular}
     \vspace{-0.5cm}
\end{table*}

\subsection{Experiment-3: Long Corridor}

In Experiment-3, we used Hokuyo LiDARs, 
which generates 3D point clouds with fewer densities.
Therefore, our proposed FV-based re-id is not adequate. 
Instead, we have used a simpler feature as an alternate of $P_1$, 
\textit{i.e.,} the heights of pedestrians. 
As shown in Fig. \ref{fig:5fmap}, \ref{fig:5fphoto},  7 Hokuyo LiDARs are installed.
Since the scan areas of neighboring LiDARs are overlapped, similarly with Experiments-1/-2,  we eliminated some areas (the three areas by blue rectangles) for this experiment.

We have collected the data from 11 am, Jan. 20th, 2022 till 1 pm, Jan. 24th, 2022.  The total number of trajectories was 15,101, 
and break down is, Jan 20th (Thu.): 3,691, 
Jan 21st (Fri.): 3,901, 
Jan 22nd (Sat.): 437, 
Jan 23rd (Sun.): 5,864, 
and Jan 24th (Mon.):1,208.

\begin{figure}[t]
    \centering
    \includegraphics[width=0.28\textwidth]{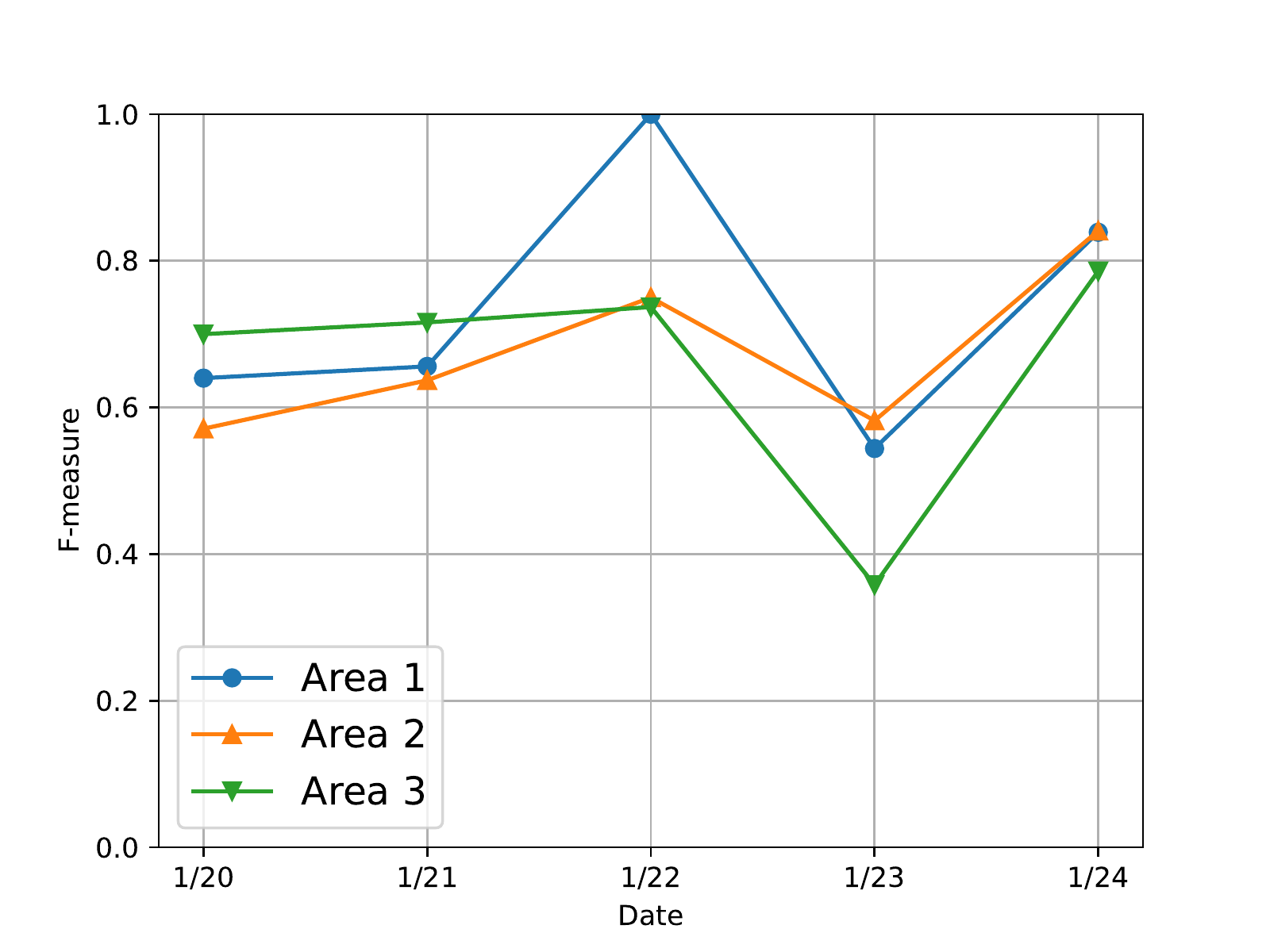}
    \vspace*{-0.3cm}
    {\caption{Accuracy with Hourly Update}}
    \label{fig:5f_res}
    \vspace{-0.4cm}
\end{figure}
We have measured the matching accuracy with hourly updates of distributions. The change of F-measure is shown in Fig. \ref{fig:5f_res}. 
The accuracy on the 3rd day (\textit{i.e.,} estimated with the previous two days' distributions) was high, and that in the 4th day (\textit{i.e.,}, estimated with unusual trajectories on Saturday) decreased. 
Except for those cases,  the accuracy increased with updated distributions; on the 5th day, it was 0.80. 

\section{Conclusion}

This paper proposed a privacy-preserving pedestrian tracking system using multiple distributed LiDARs of non-overlapping views. 
We deployed the system in a large-scale testbed with 70 colorless LiDARs and conducted three different experiments. The evaluation result on 32 participants confirms the system's ability to accurately track the pedestrians with a 0.98 F-measure even with zero-covered areas.
\vspace{-0.1cm}

\section*
{Acknowledgment}
{
  This work was supported by MEXT ``Innovation Platform for Society 5.0'' Program Grant Number JPMXP0518071489. 
T his work was partially supported by JSPS KAKENHI Grant number 22K12011 and by NVIDIA award. 
}
  
\bibliographystyle{unsrt}
\bibliography{reference}

\end{document}

%% file: definition.tex
\def\sthree#1{\mathbb{S}^3_{#1}}
\def\stwo#1{\mathbb{S}^2_{#1}}

% time, lidar, j-th segment
\def\seg#1#2#3{h^{#1}_{#2,#3}}
% set of segments, 
\def\segset#1#2{H^{#1}_{#2}}

% set of trajectories at time #1 in Lidar #2
\def\traj#1#2#3{tr^{#1}_{#2,#3}} 
\def\trajset#1#2{TR^{#1}_{#2}} 

\def\lidarset{L}

\def\t#1{t_{#1}}

\def\aff#1#2{A({#1},{#2})}

\def\trans{Q}

\def\timedist#1{p_{time}(#1)}